\documentclass[lettersize,journal]{IEEEtran}

\IEEEoverridecommandlockouts                              % This command is only needed if 
\usepackage{lipsum}
\usepackage{colortbl}
\usepackage{url}
\usepackage{graphicx}
\usepackage{amsmath}
\usepackage{amssymb}
\usepackage{multirow}
\usepackage{array}
\usepackage[linesnumbered,ruled,vlined]{algorithm2e}
\usepackage{booktabs}
\newtheorem{definition}{Definition}[section]
\usepackage[noadjust]{cite}
\usepackage{soul}
\usepackage[colorlinks,allcolors=black]{hyperref}

\ifCLASSINFOpdf
\else
\fi
\ifCLASSOPTIONcompsoc
\else
  \usepackage[caption=false,font=footnotesize]{subfig}
\fi
\begin{document}

% paper title
% Titles are generally capitalized except for words such as a, an, and, as,
% at, but, by, for, in, nor, of, on, or, the, to and up, which are usually
% not capitalized unless they are the first or last word of the title.
% Linebreaks \\ can be used within to get better formatting as desired.
% Do not put math or special symbols in the title.
\title{Dimension-variable Mapless Navigation with Deep Reinforcement Learning}
%
%
% author names and IEEE memberships
% note positions of commas and nonbreaking spaces ( ~ ) LaTeX will not break
% a structure at a ~ so this keeps an author's name from being broken across
% two lines.
% use \thanks{} to gain access to the first footnote area
% a separate \thanks must be used for each paragraph as LaTeX2e's \thanks
% was not built to handle multiple paragraphs
%

% \author{Michael~Shell,~\IEEEmembership{Member,~IEEE,}
%         John~Doe,~\IEEEmembership{Fellow,~OSA,}
%         and~Jane~Doe,~\IEEEmembership{Life~Fellow,~IEEE}% <-this % stops a space
% \thanks{M. Shell was with the Department
% of Electrical and Computer Engineering, Georgia Institute of Technology, Atlanta,
% GA, 30332 USA e-mail: (see http://www.michaelshell.org/contact.html).}% <-this % stops a space
% \thanks{J. Doe and J. Doe are with Anonymous University.}% <-this % stops a space
% \thanks{Manuscript received April 19, 2005; revised August 26, 2015.}}
\author{Wei Zhang, Yunfeng Zhang, Ning Liu and Kai Ren
\thanks{Wei Zhang and Yunfeng Zhang are with the Department of Mechanical Engineering, National University of Singapore.
	{ e-mail: weizhang@u.nus.edu, mpezyf@nus.edu.sg}
	}%
\thanks{Ning Liu are with the Smart Manufacturing Division, Advanced Remanufacturing and Technology Centre, Singapore.
	{ e-mail: Liu\_Ning@artc.a-star.edu.sg}
	}%
\thanks{Kai Ren are with the School of Mechanical Engineering, Zhejiang University, China.
	{ e-mail: ren\_kai@zju.edu.cn}
	}%
}
\maketitle
% As a general rule, do not put math, special symbols or citations
% in the abstract or keywords.
\begin{abstract}
Deep reinforcement learning (DRL) has exhibited considerable promise in the training of control agents for mapless robot navigation. However, DRL-trained agents are limited to the specific robot dimensions used during training, hindering their applicability when the robot's dimension changes for task-specific requirements. To overcome this limitation, we propose a dimension-variable robot navigation method based on DRL. Our approach involves training a meta agent in simulation and subsequently transferring the meta skill to a dimension-varied robot using a technique called dimension-variable skill transfer (DVST). During the training phase, the meta agent for the meta robot learns self-navigation skills with DRL. In the skill-transfer phase, observations from the dimension-varied robot are scaled and transferred to the meta agent, and the resulting control policy is scaled back to the dimension-varied robot. Through extensive simulated and real-world experiments, we demonstrated that the dimension-varied robots could successfully navigate in unknown and dynamic environments without any retraining. The results show that our work substantially expands the applicability of DRL-based navigation methods, enabling them to be used on robots with different dimensions without the limitation of a fixed dimension. The video of our experiments can be found in the supplementary file.

\end{abstract}

% Note that keywords are not normally used for peerreview papers.
% \begin{IEEEkeywords}
% IEEE, IEEEtran, journal, \LaTeX, paper, template.
% \end{IEEEkeywords}

% For peer review papers, you can put extra information on the cover
% page as needed:
% \ifCLASSOPTIONpeerreview
% \begin{center} \bfseries EDICS Category: 3-BBND \end{center}
% \fi
%
% For peerreview papers, this IEEEtran command inserts a page break and
% creates the second title. It will be ignored for other modes.
\IEEEpeerreviewmaketitle

\section{Introduction}
% The very first letter is a 2 line initial drop letter followed
% by the rest of the first word in caps.
% 
% form to use if the first word consists of a single letter:
% \IEEEPARstart{A}{demo} file is ....
% 
% form to use if you need the single drop letter followed by
% normal text (unknown if ever used by IEEE):
% \IEEEPARstart{A}{}demo file is ....
% 
% Some journals put the first two words in caps:
% \IEEEPARstart{T}{his demo} file is ....
% 
% Here we have the typical use of a "T" for an initial drop letter
% and "HIS" in caps to complete the first word.
\IEEEPARstart{T}{he} ability of reaching a specified destination while avoiding obstacles, known as self-navigation, is a fundamental skill for mobile robots. Map-based approaches \cite{Yue2021col,Zou2022A} require the robot to determine its location, estimate its current state, identify obstacles, and plan both global and local paths. However, these methods are computationally intensive and  less effective in dynamic, unknown, or unstructured scenarios such as search and rescue tasks. Recently, deep reinforcement learning (DRL) \cite{mnih2015human} has emerged as a promising solution for mapless navigation, demonstrating notable achievements \cite{Wang2023,Li2022MSN}. By leveraging DRL, a mobile robot can learn to navigate to its destination through interaction with the training environment, using deep neural networks (DNNs) \cite{lecun2015deep} to directly generate motion commands from raw observations. The trained DRL agent can navigate in dynamic environments efficiently \cite{Zhou2022Navigating}. Moreover, the commands generated from the trained DNNs are computationally efficient \cite{Lim2020}. Due to these characteristics, DRL-based methods have gained considerable attention in the field of robot navigation \cite{Wu2022}.

To mitigate the substantial cost associated with real-world training, the employment of the sim-to-real (simulation-to-reality) approach has become a widely adopted strategy for learning navigation controllers using DRL. The basic procedure for sim-to-real robot navigation involves training the DRL agent in a simulated environment and subsequently deploying the acquired controller to a physical robot for real-world tasks \cite{Tai2017}. In order to enhance the DRL agent's performance in real scenarios, Shi et al. \cite{shi2020} utilized the curiosity-based reward to encourage the robot's exploration of unexplored areas. In addition, Xie et al. \cite{xie2021} employed conventional controllers to expedite the training process of the DRL agent. Similarly, Rana et al. \cite{Rana2023} utilized both conventional controllers and an uncertainty-aware training strategy to accelerate the training of the DRL agent. Furthermore, to address the challenges posed by local minimum issues in mapless navigation, Jang et al. \cite{Jang2022hind} introduced hindsight intermediate targets for the DRL agent only using recent observations. It is noteworthy that some of the DRL-based navigation controllers, despite being trained using 2D input in simulated scenarios, can be effectively utilized for robot navigation in 3D environments through the incorporation of depth sensors, such as a depth camera. This adaptation does not require any additional training, as implemented in \cite{Leiva2020}.

Currently, the primary focus of studies on DRL-based robot navigation revolves around enhancing the navigation performance of controllers designed for robots with fixed dimensions \cite{shi2020,Pfeiffer2018,xie2021,Jang2022hind,Wu2022,Lim2020,Tai2017,Rana2023,Leiva2020}. However, in practical applications, the dimensions of robots may vary due to task requirement. For instance, when a robot is tasked with transporting a cargo or a victim that exceeds its physical dimensions, utilizing the default DRL controller under such circumstances leads to inevitable task failures. To address this issue, retraining the DRL agent using the dimension-scaled robot model is the simplest approach. However, the retraining process is time-consuming, which poses a significant cost to real-world applications. In comparison, skill transfer presents a more viable alternative. Specifically, skill transfer involves transferring the previously acquired navigation skills to dimension-varied robots. One commonly employed technique for skill transfer is meta learning \cite{yu2020,Ali2021Bayesian}, which enables the robot to adapt to unforeseen circumstances through online learning. However, these methods necessitate the acquisition of several tasks initially, still requiring some time for real-world retraining, and imposing high computational demands on the on-board processor. Additionally, transfer learning using domain randomization \cite{Antonio2020} is another skill-transfer method; however, it primarily focuses on transferring learned skills from the training environment to the same robot operating in a new environment. In sharp contrast, our research problem focuses on transferring the learned skills to robots with varied dimensions.

In this paper, we propose a general framework for a single DRL-based controller to work on mobile robots with changed dimensions. Our approach first trains a meta robot in simulation, enabling the meta agent to acquire high-performance navigation skills through the proposed meta-skill learning (MSL) algorithm. Subsequently, when there is a change in the robot dimension, the meta robot can directly transfer its navigation skill (inside the meta agent) to the dimension-varied robot utilizing the proposed dimension-variable skill transfer (DVST) method. Hence, we referred to our method as MSL\_DVST. In summary, the key contributions of this paper are outlined as follows:

\begin{enumerate}[]
	\item The dimension-variable robot navigation method has been proposed to extend the applicability of DRL-based controllers beyond the fixed dimension to encompass varied dimensions.
	\item The DVST method has been proposed, enabling the direct transfer of the DNN controller, trained within a simulation environment for the meta robot, to a dimension-scaled robot without any retraining. Moreover, DVST is not limited to navigation agents trained through MSL, it is also compatible with other DNN-controlled navigation agents that utilize depth sensors for perception..
	\item Extensive real-world experiments have been conducted and the testing results have validated the effectiveness of our MSL\_DVST approach.

\end{enumerate}

The rest of this paper is organized as follows. A brief introduction to the dimension-variable robot navigation problem is given in Section \ref{Prel}. The proposed MSL\_DVST method is described in Section \ref{approach}, followed by simulation and real-world experiments and the corresponding results in Section \ref{implementation}. Last, we draw the conclusions in Section \ref{conclusion}.

\section{Preliminaries}\label{Prel}
In this work, we aim to train a DNN as a real-time navigation controller for a mobile robot and its dimension-varied versions. Specifically, given the dimension-varied robot, the DNN controller can drive the robot to its goal safely without retraining. The abbreviations of terms frequently used  in this paper are listed in Table \ref{abbreviations}.
\begin{table}
\caption{\label{abbreviations} The list of abbreviations.} 
\renewcommand\arraystretch{1.2}
\centering
% Please add the following required packages to your document preamble:
% \usepackage{multirow}
\setlength{\tabcolsep}{1.4mm}{\begin{tabular}{ll}
\hline
\hline
IP Methods   & Descriptions                                                 \\ \hline
DNN            & Deep neural network    \\
DRL            & Deep reinforcement learning    \\
DSR            & Dimension-scaled robot    \\
DVR            & Dimension-varied robot  \\
DVST         & Dimension-variable Skill Transfer \\ 
MR              & Meta robot             \\
MSL          & Meta-skill learning \\
SAC            & Soft actor-critic    \\
\hline
\hline
\end{tabular}}
\end{table}

\subsection{Problem Description}\label{prob_desc}

\begin{figure}[t]
    \centering
	  \subfloat[]{
       \includegraphics[width=0.45\linewidth]{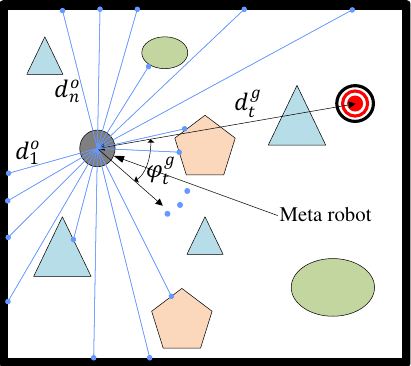}\label{problem1}}
	  \subfloat[]{
        \includegraphics[width=0.45\linewidth]{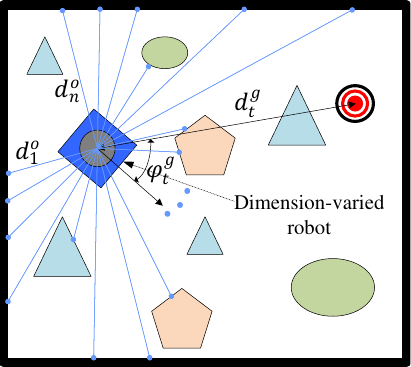}\label{problem2}}
	\caption{Illustration of the mapless robot navigation problem for (a) Meta robot and (b) dimension-varied robot.}
	\label{problem}
\end{figure}

The mapless robot navigation problem can be conceptualized as a sequential decision-making process. As depicted in Fig. \ref{problem1}, a meta robot is required to reach the goal position while avoiding collisions with obstacles. To perceive its surroundings, the robot is equipped with depth sensors on the top center. This paper adopts 2D LiDAR as the depth sensor. For studies employing 3D depth sensors, such as a depth camera, the integration of 3D depth data into the 2D LiDAR representation can be achieved through the method introduced by \cite{Leiva2020}. We assume the relative position of goal in robot coordinate frame $s_t^g=\{d_t^g,\varphi_t^g\}$ can be determined through localization techniques such as sound source localization \cite{Ishfaque2023} or WIFI localization \cite{Ding2022}. We denote the input of the DNN controller as $s_t=\{s_t^o,s_t^g,v_{t},\omega_{t}\}$, where $s_t^o$ corresponds to the readings from the distance sensors. The action $a_t$ of the robot  is the velocity command. Given $s_t$, the robot executes action $a_t$ under current policy $\pi$. Subsequently, the robot updates its input to $s_{t+1}$ based on new observations and receives a reward \( r_{t}\) calculated by the reward function. The specific form of the reward function is as follows,
\begin{equation}
r_t = 
\begin{cases}
r_{s}, & \text{if success}\\
r_{c}, & \text{if collision}\\
c_{1} \left( d_{t}^{g}-d_{t+1}^{g} \right) -c_{2}, & \text{otherwise}.
\end{cases}
\end{equation}
where $c_1$ is a scaling constant, $c_{2}$ is a small constant working as a time penalty. The reward function comprises three distinct components: a positive component $r_s$ that incentivizes successful outcomes, a negative component $r_c$ that penalizes collisions, and a small dense component that encourages the robot to move closer to the target. The objective of this decision-making process is to determine an optimal policy denoted as $\pi^\ast$, which maximizes the cumulative discounted rewards $G_{t} = \sum_{\tau=t}^{T} \gamma^{\tau-t}r_{\tau}$, where $\gamma \in [0,1]$ represents the discount factor.

The dimension-variable mapless navigation problem is an extension of the mapless navigation problem. As shown in Fig. \ref{problem2},  when the meta robot carries a cargo or victim, its dimensions increase, rendering the existing controller ineffective for the dimension-varied robot. Therefore, the aim of this study is to build a controller, using the meta-robot controller as a basis, that enables the dimension-varied robot to perform navigation tasks at a performance level comparable to the meta robot.

\subsection{Soft Actor Critic}
In this study, Soft Actor-Critic (SAC) \cite{Haarnoja2018} has been selected as the DRL algorithm for learning navigation skills, primarily due to its remarkable performance in DRL-based robot navigation \cite{de2021soft}. The SAC algorithm seeks to maximize the expected cumulative reward, taking into account entropy regularization. Specifically, in SAC, the Q-function \(Q^{ \pi } \left(s,a \right) \)  represents the expected return achieved by taking action \(a\) given input \( s \),
\begin{equation}
\begin{small}
\begin{aligned}
Q^{\pi}\left(s,a\right)=\mathbb{E}_{\pi}\left[G_{t=0}+\alpha\sum_{t=0}^{T}\gamma^{t}H\left(\pi \left(\cdot\vert s_{t}\right)\right)\vert s_{0}=s,a_{0}=a \right],
\end{aligned}
\end{small}
\end{equation}
where the entropy term \(H\left(\pi\left(\cdot\vert s_{t}\right)\right) =- \int _{\vert A \vert }^{} \pi  \left( a \vert s_{t} \right) \log  \pi  \left( a \vert s_{t} \right) da \)  measures the uncertainty associated with the action selection process. Additionally, the factor \(\alpha >0\) governs the trade-off between the contributions of the discounted cumulative reward and the entropy term. In SAC, two critic (\textit{Q}) networks (parameterized by  \(  \phi _{1} \)  and  \(  \phi _{2} \)) are employed to estimate the \textit{Q} value. By drawing a mini-batch $\mathcal{M}$ from the replay buffer, the loss function $\mathcal{L}(\phi_{1,2})$ for the Q function can be expressed as follows,
\begin{equation}
\begin{small}
\begin{aligned}
&\mathcal{L}(\phi_{1,2})=\frac{1}{\left|\mathcal{M}\right|}\ \sum_{\left(s,a,r,s^\prime,d\right)\in\mathcal{M}}\left(Q\left(s,a|{\phi}_{1,2}\right)-\hat{Q}(s,a)\right)^2,\\
&\hat{Q}(s,a)=r+\gamma(1-d))(\min_{j=1,2}Q(s^\prime,\widetilde{a}^\prime|{\hat{\phi}}_{j})-\alpha\log\pi_{\theta}(\widetilde{a}^\prime\vert s^\prime)),
\end{aligned}
\end{small}
\end{equation}
where $\hat{Q}(s,a)$ represents the target Q value, and $\widetilde{a}^\prime$  is obtained by sampling from the action distribution $\pi\left(\cdot\vert s^\prime\right)$. The objective of optimizing the policy network is to minimize the following loss function $\mathcal{L}(\theta)$, which is defined as follows, 
\begin{equation}
\begin{small}
\begin{aligned}
\mathcal{L}(\theta)=\frac{1}{\left|\mathcal{M}\right|}\ \sum_{x\in\mathcal{M}}\left(\alpha\log{\pi(\widetilde{a}\vert s,\theta)-\min_{j=1,2}Q(s,\widetilde{a}|{\phi}_{j})}\right).
\end{aligned}
\end{small}
\end{equation}

\section{APPROACH}\label{approach}
\begin{figure}[!t]
	\centering
	\includegraphics[width=0.95\linewidth]{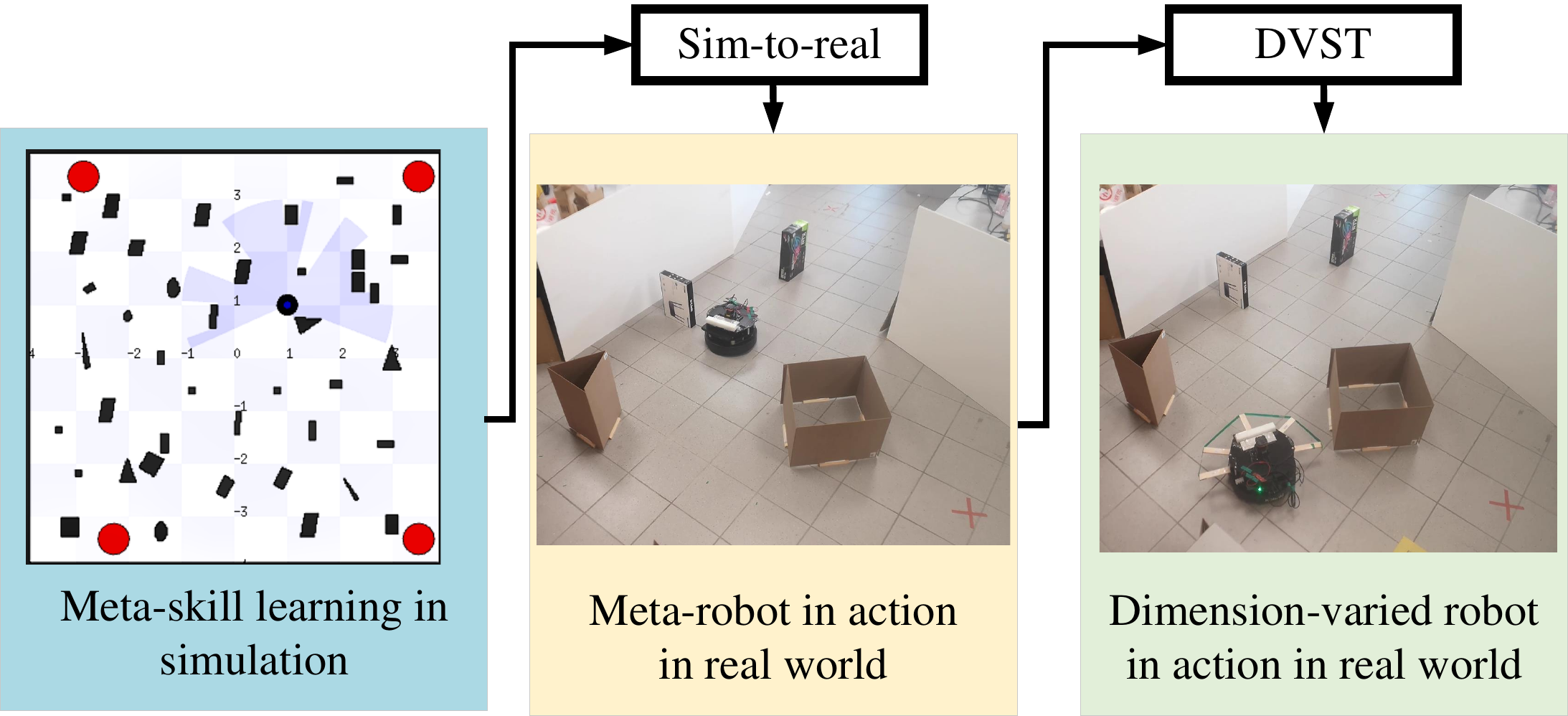}
	\caption{The overall framework of the proposed method.}
	\label{fig:2}
\end{figure}

In this section, we propose a new method named MSL\_DVST that allows a mobile robot to adaptively adjust its control strategies when its dimension changes. The overall framework of this method is given in Fig. \ref{fig:2}. As shown, it contains two stages, i.e., meta-skill learning and DVST. In the first stage, a meta agent is trained using a meta robot model in well-designed simulation environments to master navigation skills in crowded scenarios. In the second stage, the DVST method is utilized for transferring the meta-skill to robots with changed dimensions. The primary focus of this paper is to transfer the navigation skills acquired by the meta robot to dimension-varied robots. It should be emphasized that the presented DVST method is applicable to a wide range of DNN-controlled robots and is not limited solely to the meta agent used in this study. Generally, the efficacy of the skill transfer process is influenced by the performance of the meta agent, as a higher-performing meta agent typically leads to improved performance in the skill-transferred agent.\par
\subsection{Meta-skill Learning}

\begin{figure}[t]
    \centering
	  \subfloat[]{
       \includegraphics[width=0.45\linewidth]{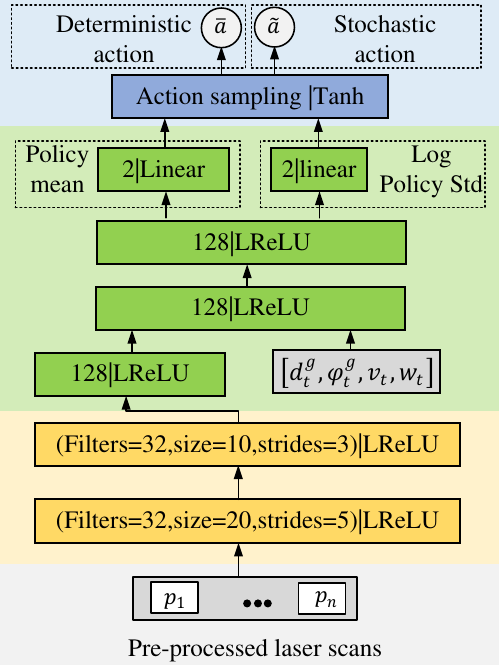}\label{net1}}
	  \subfloat[]{
        \includegraphics[width=0.45\linewidth]{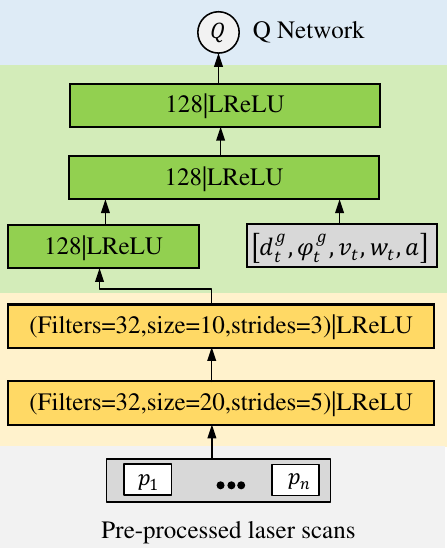}\label{net2}}
	\caption{Neural network structures used in meta-skill learning. (a) Policy (actor) network; (b) Critic (Q) network.}
	\label{networks}
\end{figure}

The learning objective of the meta agent is to master navigation skills in crowded scenarios. To achieve this learning objective, we build our meta-skill learning algorithm on SAC, which is found to have better performance in robot navigation than other model-free DRL methods \cite{de2021soft}, such as deep deterministic policy gradient (DDPG) \cite{Lillicrap2016}. The employed neural network structure of SAC is given in Fig. \ref{networks}, consisting of a policy network parameterized by $\theta$ and two  critic networks (sharing the same network structure) parameterized by  \(  \phi _{1} \)  and  \(  \phi _{2} \). For each network,  \( n \)  laser beams ( \( n=540 \) in this paper) are pre-processed by reciprocal function \cite{zhang2022ipaprec} and fed into 1D CNN layers for feature extraction. The extracted features are flattened and concatenated with robot velocities and the goal position (and the action for critic network). The combined features are then fed into fully-connected (FC) layers for computing the action and Q-values. In addition, policy regularization \cite{liu2021regularization}, specifically $L_{2}$ regularization, is employed to improve the generalization performance of the meta agent.

The meta agent is expected to master the navigation skill in very crowded scenarios. Directly training the navigation agent in a crowded scenario is challenging, partly due to the high collision rate and sparse goal-reaching rate. To address this problem, we follow the curriculum learning (CL) paradigm \cite{Bengio2009}: start from simple navigation tasks and gradually increase the task difficulty. For robot navigation, the task difficulty is mainly determined by the crowded level of the navigation scenario: the more crowded the scenario, the more complex the navigation task is. In this paper, all the training scenarios in the training curriculum share the same layout but with different room sizes. The scenarios in the curriculum are ordered by their room size: the room size of the next training scenario must be shorter than that of the current one. By reducing the room size, the subsequent task is guaranteed to be harder than the current one. Considering the exploration noise in DRL, the criterion of task completion in each scenario is that the agent reaches a success rate of $90\%$ in the recent 50 episodes. After finishing one task, the agent will be transited to the next scenario according to the curriculum. The training will end when the total training steps exceed the pre-defined value. The detailed description of MSL is given in Algorithm \ref{algorithm}.

\begin{algorithm}[t]
\label{algorithm}
\SetAlgoLined
 Initialize policy parameters $\theta$, Q-value function parameter $\phi_{1}, \phi_{2}$, empty replay buffer $\mathcal{B}$, the initial training scenario Env\_0, success rate $\rho$ \;
 \For{\upshape{episode}$=1$, 2, \ldots, }
 {\If{\upshape{$\rho=0.9$}}{select next training scenario in the curriculum;}  
 Reset the training environment and initialize $t=0$\;
      Obtain initial observation $s_0$\;   
 \While{$t<T_{max}$ \upshape{\textbf{and} not terminate}}
   {
      \eIf{\upshape{training}}{Sample action $a_t\sim\pi_{\theta}(s_t)$\;}
      {Calculate action $a_t$\ with PID controller\;}
      Execute $a_t$ in simulation\;
      Obtain next observation $s_{t+1}$, reward $r_t$, and the termination signal $d_t$\;
      Store $\{s_t,a_t,r_t,s_{t+1},d_t\}$ in $\mathcal{B}$\;
      $s_t\leftarrow s_{t+1}$, $t\leftarrow t+1$\;
      }
      		\If{\upshape{Training}}{
			Update success rate $\rho$\;
			\For{\upshape{$\tau$ in range($t$)}}
			{
				Sample a minibatch from replay buffer $\mathcal{B}$\;
				Update $\phi_{1} and \phi_{2}$\;
				\If{\upshape{$\tau$ mod 2 = 0}}
				{
					Update $\theta$\;
				}
			}
		}
      }

\caption{Meta-skill learning}
\end{algorithm}

\subsection{Dimension-variable Skill Transfer (DVST)}
\begin{figure}[!t]
	\centering
	\includegraphics[width=0.95\linewidth]{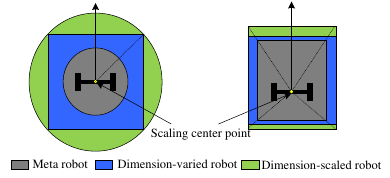}
	\caption{Illustration of the dimension-scaled robot.}
	\label{DSR}
\end{figure}

After training, the meta agent for the meta robot is ready for transferring its skill to the dimension-varied robot. For the purpose of skill transfer, a dimension-scaled robot is introduced. For simplicity, we refer to the meta robot as MR, the dimension-varied robot as DVR, and the dimension-scaled robot as DSR. The DSR is defined as having the same shape as the MR and having the smallest size that can cover the DVR. For better illustration, in Fig. \ref{DSR}, we take two MRs with circular (on the left) and rectangular (on the right) shapes as examples. As shown, both of the MRs carry square cargoes, and the resulting DVRs are marked with blue. With the MR and DVR, the DSRs are generated as follows. First, they are generated with the same shape as the MR, and the scaling center point is centered between the two drive wheels (center of axle). Second, the dimensions of them should be the smallest dimensions that can cover the DVRs. Following the two steps, the generated DSRs are generated and marked with green in Fig. \ref{DSR}. According to the definition of DSR, successful navigation by DSR implies successful navigation by DVR as well. With DSR, the key idea behind the DVST is: (1) mapping the observations of MR to the observations of DSR  (observation transfer) and (2) scaling the DRL policy generated by MR back to DSR  (policy transfer). This idea is illustrated with mathematical expressions as follows.\par

\subsubsection{Observation Transfer}
\begin{figure}[!t]
	\centering
	\includegraphics[width=0.95\linewidth]{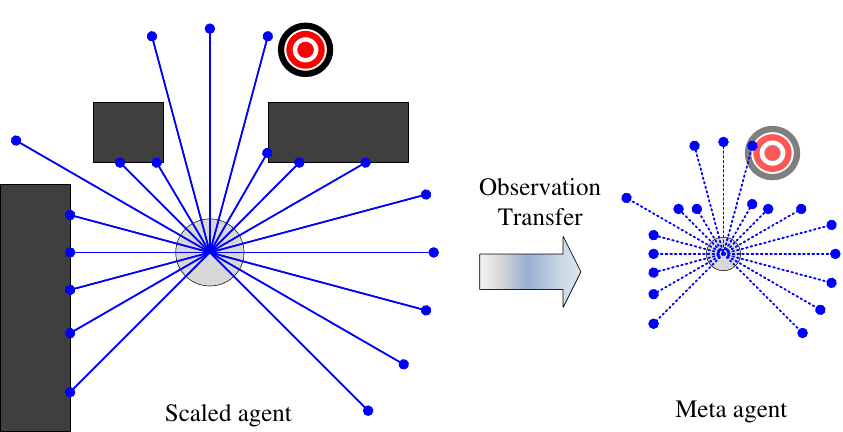}
	\caption{Illustration of the observation transfer process.}
	\label{fig:6}
\end{figure}

The observation  \( s_{s}= \left\{ s_{s}^{d},s_{s}^{di} \right\}  \)  received by  DSR  can be divided into distance-dependent observations  \( s_{s}^{d} \)  (the LiDAR readings, the relative distance between the goal and robot, and current linear velocity) and distance-independent observations  \( s_{s}^{di} \)  (relative angle between the goal and current angular velocity). The scaling ratio between the DSR and the MR is denoted as $\mu$. During the observation transfer process, as shown in Fig. \ref{fig:6}, the distance-dependent observations of  DSR  are scaled to that of  MR  based on the ratio of  \( \mu \), while the distance-independent observations remain the same. The corresponding observation representation of  \( s_{m} \)  in  MR  after observation transfer is:\par

\begin{equation}
s_{m}= \left\{ \frac{s_{s}^{d}}{\mu} ,s_{s}^{di} \right\}.
\end{equation}
\subsubsection{Policy transfer}
\begin{definition}
 For two trajectories  \( l_{k} \)  and  \( l_{j} \), if point  \( P_{k} \in l_{j} \)  holds for  \(  \forall  \)   \( P_{k} \in l_{k} \), then we define \( l_{k} \subseteq l_{j} \).
\end{definition}
\begin{figure}[!t]
	\centering
	\includegraphics[width=0.95\linewidth]{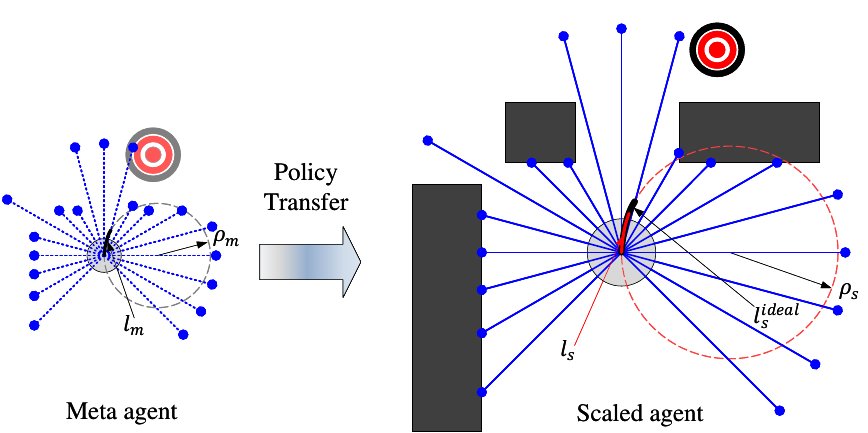}
	\caption{Illustration of the policy transfer process.}
	\label{fig:7}
\end{figure}

With  \( s_{m} \)  as input, the control policy (i.e., the command velocities) of the MR generated by the DNN-based controller can be represented as  \( v_{m} \left( \frac{s_{s}^{d}}{\mu},s_{s}^{di} \vert  \theta  \right) \)  and  \(  \omega _{m} \left( \frac{s_{s}^{d}}{\mu},s_{s}^{di} \vert  \theta  \right)  \). In the skill-transfer process, the robot is assumed to operate with constant velocities within a single control cycle, which is the same as the popular dynamic window approach \cite{Fox1997}. The resulting trajectory \( l_{m} \) within this control cycle can be represented as a circular arc (Fig. \ref{fig:7}, where the starting point corresponds to the center of axle, i.e., the origin in the robot frame. This trajectory can well approximate the real trajectory when the time interval is short and will converge to the real trajectory when the time interval goes to zero \cite{Fox1997}. The approximated trajectory can be represented by its length  \(  \vert l^{m} \vert =v_{m} \Delta T \)  and its radius  \(  \rho _{m}=\frac{v_{m}}{ \omega _{m}} \)  ( \(  \rho _{m} \)  is negative if  \(  \omega _{m} \)  is negative), where  \(  \Delta T \)  is the duration of one control cycle. To transfer the policy back to the robot  DSR, the ideal trajectory  \( l_{s}^{ideal} \)  of  DSR  should be similar to  \( l_{m} \), and the similarity ratio is  \(\frac{R_{s}}{R_{m}} \). Based on this idea, the ideal command velocities and resulting trajectory for DSR are as follows,
\begin{equation}
\begin{split}
&v_{s}^{ideal}=\mu v_{m} \left( \frac{s_{s}^{d}}{\mu},s_{s}^{di} \vert  \theta  \right),\\
&\omega _{s}^{ideal}= \omega _{m} \left( \frac{s_{s}^{d}}{\mu},s_{s}^{di} \vert  \theta  \right),\\
&\vert l_{s}^{ideal} \vert =v_{s}^{ideal} \Delta T,\\
&\rho _{s}^{ideal}=\frac{v_{s}^{ideal}}{ \omega _{s}^{ideal}}.
\end{split}
\end{equation}
However, the ideal velocities of  DSR  may exceed the velocity bounds. Constrained by velocity bounds, to make the real trajectory  \( l_{s} \)  cover the ideal trajectory as much as possible (i.e., maximize  \(  \vert l_{s} \vert  \)), the objective of the skill transfer problem can be formulated as,\par
\begin{equation}
\begin{aligned}
\mathop{\arg\max}_{\mathop{v}_{s},\mathop{ \omega }_{s}} \quad &  \vert l_{s} \vert\\
\textrm{subject to} \quad & v_{s} \leq v_{s}^{\max }, \\
&\vert  \omega _{s} \vert  \leq  \omega _{s}^{\max },    \\
&l_{s} \subseteq l_{s}^{ideal}.    \\
\end{aligned}
\end{equation}
where  \( l_{s} \subseteq l_{s}^{ideal} \)  is equivalent to  \(  \rho _{s}= \rho _{s}^{ideal} \)  and  \(  \vert l_{s} \vert \leq \vert l_{s}^{ideal} \vert  \) ( \( l_{s} \)  and  \( l_{s}^{ideal} \)  share the same start point, i.e., the original point in robot frame). Hence, the above optimization problem can be rewritten as:\par

\begin{equation}
\begin{aligned}
\mathop{\arg\max}_{\mathop{v}_{s},\mathop{ \omega }_{s}} \quad &  v_{s} \Delta T\\
\textrm{subject to} \quad & v_{s} \leq \min  \left\{ v_{s}^{\max },v_{s}^{ideal} \right\}, \\
&\vert  \omega _{s} \vert  \leq  \omega _{s}^{\max },    \\
&\frac{v_{s}}{ \omega _{s}}= \rho _{s}^{ideal}.    \\
\end{aligned}
\end{equation}
The solution to the above problem is piecewise conditioned on whether the ideal radius of curvature  \(  \rho _{s}^{ideal} \)  can be achieved with the maximum linear velocity  \( v_{s}^{\max } \)  or not. If  \( \frac{v_{s}^{\max }}{ \omega _{s}^{\max }} \leq  \vert  \rho _{s}^{ideal} \vert  \), the ideal radius of curvature  \(  \rho _{s}^{ideal} \)  can be achieved with maximum linear velocity  \( v_{s}^{\max } \)  by adjusting the angular velocity  \( w_{s} \). In other words, given any linear velocity  \( v_{s} \)  within the velocity bound, we can always find an angular velocity  \(  \omega _{s} \)  within the velocity bound that can satisfy  \( \frac{v_{s}}{ \omega _{s}}= \rho _{s}^{ideal}. \)  Therefore, under this condition, we need to calculate  \( v_{s} \)  first. To maximize  \( v_{s} \), constrained by  \( v_{s} \leq \min  \left\{ v_{s}^{\max },v_{s}^{ideal} \right\}  \),  \( v_{s} \)  should be the smaller value of  \( v_{s}^{\max } \)  and  \( v_{s}^{ideal} \). Hence, the velocities of robot  DSR  are as follows,\par
\begin{equation}
\label{eq:vel1}
\begin{aligned}
v_{s}&=\min  \left\{ v_{s}^{ideal},v_{s}^{\max } \right\},\\
\omega _{s}&=\frac{v_{s}}{ \rho _{s}^{ideal}}. \\
\end{aligned}
\end{equation}
Else (i.e.,  \( \frac{v_{s}^{\max }}{ \omega _{s}^{\max }}> \vert  \rho _{s}^{ideal} \vert  \)), the ideal radius of curvature  \(  \rho _{s}^{ideal} \)  cannot be achieved with maximum linear velocity  \( v_{s}^{\max } \). In other words, given an angular velocity  \(  \omega _{s} \)  within the velocity bound, we can always find a linear velocity  \( v_{s} \)  within the velocity bound that can ensure  \( \frac{v_{s}}{ \omega _{s}}= \rho _{s}^{ideal} \). Therefore, under this condition, we need to calculate  \(  \omega _{s} \)  first. As  \( v_{s} \)  is proportional to  \(  \vert  \omega _{s} \vert  \), maximizing  \( v_{s} \)  is equivalent to maximizing  \(  \vert  \omega _{s} \vert  \). To maximize  \(  \vert  \omega _{s} \vert  \), constrained by  \(  \vert  \omega _{s} \vert  \leq \min  \left\{  \omega _{s}^{\max }, \vert  \omega _{s}^{ideal} \vert  \right\}  \),  \(  \vert  \omega _{s} \vert  \)  should be the smaller value of  \(  \omega _{s}^{\max } \)  and  \(  \vert  \omega _{s}^{ideal} \vert  \). Hence, the velocities of robot  DSR  are as follows,\par
\begin{equation}
\label{eq:vel2}
\begin{aligned}
\omega _{s}&=\textrm{Sgn} \left(  \omega _{s}^{ideal} \right) \min  \left\{  \omega _{s}^{\max }, \vert  \omega _{s}^{ideal} \vert  \right\} ,\\
v_{s}&= \omega _{s} \rho _{s}^{ideal}.
\end{aligned}
\end{equation}
Based on (\ref{eq:vel1}) and (\ref{eq:vel2}), the meta-skill can be transferred to the dimension-scaled robot and the corresponding dimension-varied robot.\par
\section{Implementation and Tests in Simulation}\label{implementation}
In this section, we used a circular meta-robot as an example to illustrate the training process in MSL. Subsequently, the learned meta-skills were successfully transferred to different dimension-varied robots to perform navigation tasks.

\subsection{Meta-skill Training}
\begin{figure}[t]
    \centering
	  \subfloat[]{
        \includegraphics[width=0.292\linewidth]{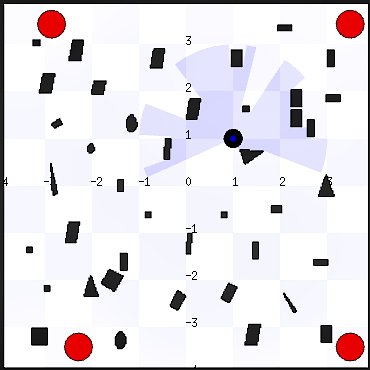}\label{train1}}
	  \subfloat[]{
        \includegraphics[width=0.255\linewidth]{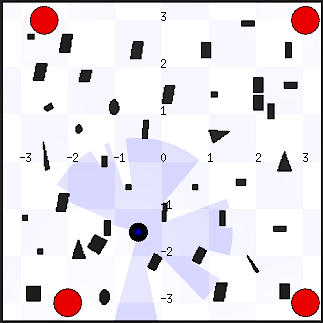}\label{train2}}
        \subfloat[]{
        \includegraphics[width=0.219\linewidth]{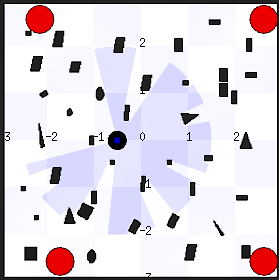}\label{train3}}
    	\subfloat[]{
        \includegraphics[width=0.183\linewidth]{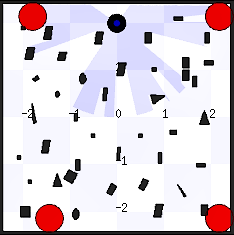}\label{train4}}
	\caption{Training environments for meta agent learning. (a) Env\_0 ($8\times8\text{m}^2$); (b) Env\_1 ($7\times7\text{m}^2$); (c) Env\_2 ($6\times6\text{m}^2$); (d) Env\_3 ($5\times5\text{m}^2$)}.
	\label{sim_env}
\end{figure}

The training process was conducted within the \textit{ROS Stage} simulator \cite{vaughan2008massively}, which is a lightweight simulator designed for mobile robots. As shown in Fig. \ref{sim_env}, the initial training scenario was Env\_0 (Fig. \ref{train1}), an $8\times8\text{m}^2$ room which contained various shapes of obstacles and much free space. In the learning curriculum, the room side length of the next training scenario was one meter shorter than that of the current one. Initially, the meta robot (represented in black in Fig. \ref{sim_env}) and its target (not visualized) were randomly positioned in obstacle-free regions. The meta robot had a radius of 0.2m, with linear and angular velocity constraints of 0.5m/s and $\frac{\pi}{2}$ rad/s, respectively. It was equipped with a LiDAR sensor that possessed a field of view of $270^\circ$, providing an angular resolution of $0.5^\circ$. 

\begin{figure}[!t]
	\centering
	\includegraphics[width=0.95\linewidth]{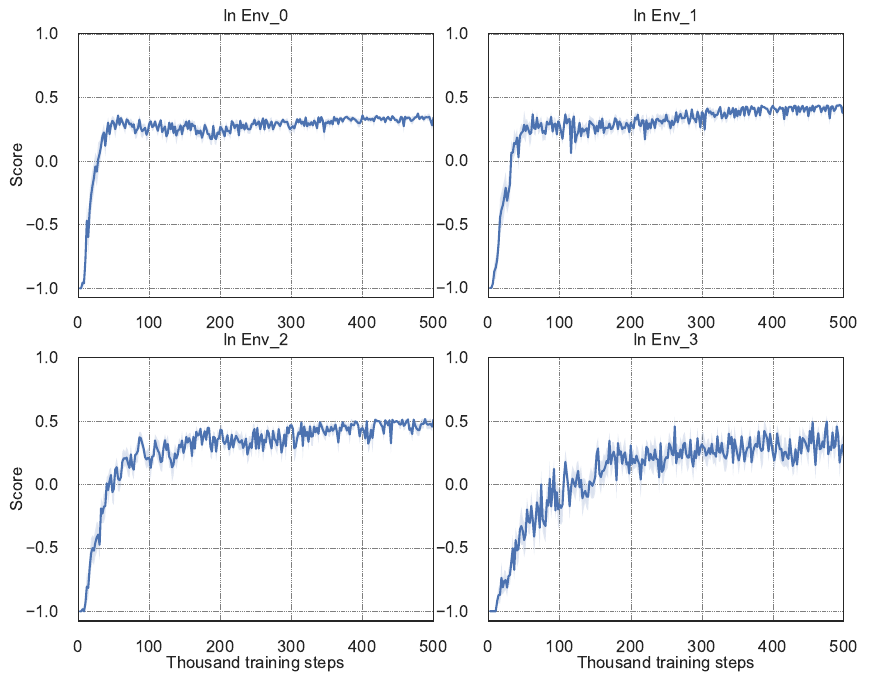}
	\caption{Average scores obtained by the meta agent in Env\_0-Env\_3. The translucent areas represent the standard deviation of the score.}
	\label{learning_curve}
\end{figure}

During the initial 100 episodes, the meta agent employed actions generated by a PID controller \cite{xie2021}. Subsequently, it switched to the stochastic policy following a squashed Gaussian distribution. Each episode was terminated under three conditions: when the robot successfully reached the goal, when it collided with obstacles, or when the maximum allowable number of steps was exceeded. The training process spanned a duration of $5\times10^5$ steps, with the trained agent being assessed every $2\times10^3$ steps. In the evaluation phase, in each training scenario, the robot was tasked with sequentially reaching four goal points (indicated by red points in Fig. \ref{sim_env}) from the original point. To evaluate the stability and repeatability of the learning algorithm, the training process was repeated ten times using different random seeds. Consistent with \cite{zhang2022ipaprec}, the navigation task metric employed in this study was based on the robot's terminal state and the time required to reach the destination. This metric was quantified by the score function $S$
, defined as follows, 
\begin{equation}
S = 
\begin{cases}
-1, & \text{if collision or time out}\\
1-2\frac{T_s}{T_{max}}, & \text{if success}.
\end{cases}
\end{equation}
where $T_s$ represented the number of navigation steps taken by the agent, while $T_{max}$ ($T_{max}=400$ in this paper) denoted the predetermined maximum allowed steps. This metric prevented early collision cases from receiving higher scores compared to the instances of collisions that occurred later in the navigation process. Among the ten-time training process, all the meta agents ended their training in Env\_3 with an average success rate of 81.4\%. The learning curves depicting the progress of the meta agent in the four testing scenarios were plotted in Fig. \ref{learning_curve}. As depicted, the meta agent progressively acquired the capability to navigate within crowded scenarios. 

\par

\subsection{Performance Evaluation of MSL\_DVST In Simulation}
\begin{figure}[!t]
	\centering
	\includegraphics[width=0.95\linewidth]{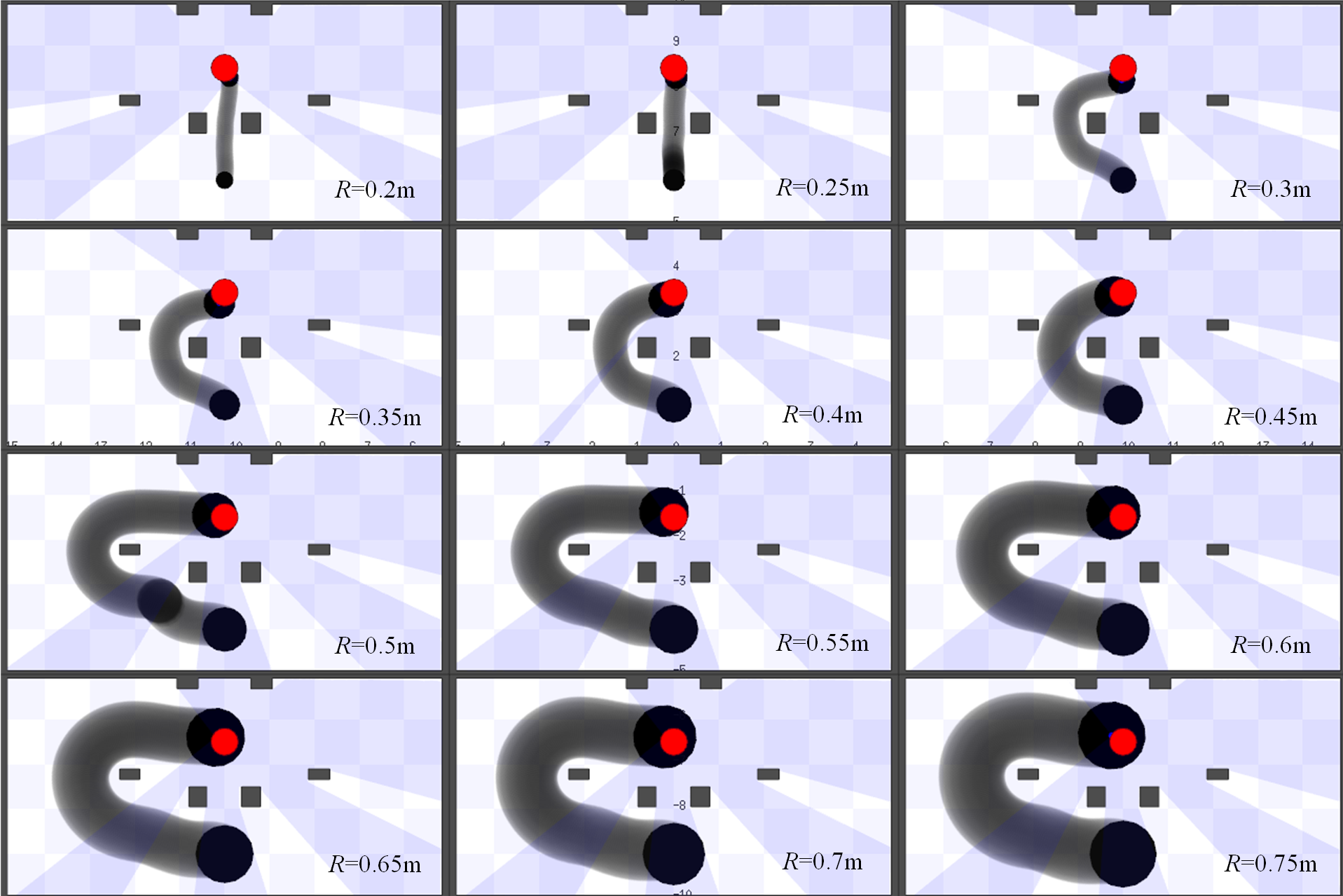}
	\caption{Trajectories of dimension-varied robots in the testing scenario, utilizing MSL\_DVST as the controller.}
	\label{sim_test}
\end{figure}

To investigate the performance of the navigation controller provided by MSL\_DVST, we tested 12 robots with different dimensions in a simulated scenario shown in Fig. \ref{sim_test}. As shown, there are 12 identical testing rooms, and each room had one dimension-varied circular robot. The radii of tested robots  $R$ (labelled in Fig. \ref{sim_test}) ranged from 0.2m to 0.75m with an interval of 0.05m. In each room, with the room center as the origin, the robot started at coordinates (0, -1.5), while its goal was positioned at coordinates (0, 1.5). After testing, the resulting trajectories of the robots were also plotted in Fig. \ref{sim_test}. As shown, with relatively small radii, i.e., 0.2m to 0.25m, the robot efficiently traversed the obstacle gate, taking a nearly optimal straight-line path to the target. When \textit{R} reached 0.3m and beyond, the robot intelligently recognized its inability to pass the front gate and turned to the left path instead. The testing results demonstrated the robot's capacity to dynamically select appropriate paths based on its dimension. \par
\begin{figure}[!t]
	\centering
	\includegraphics[width=0.95\linewidth]{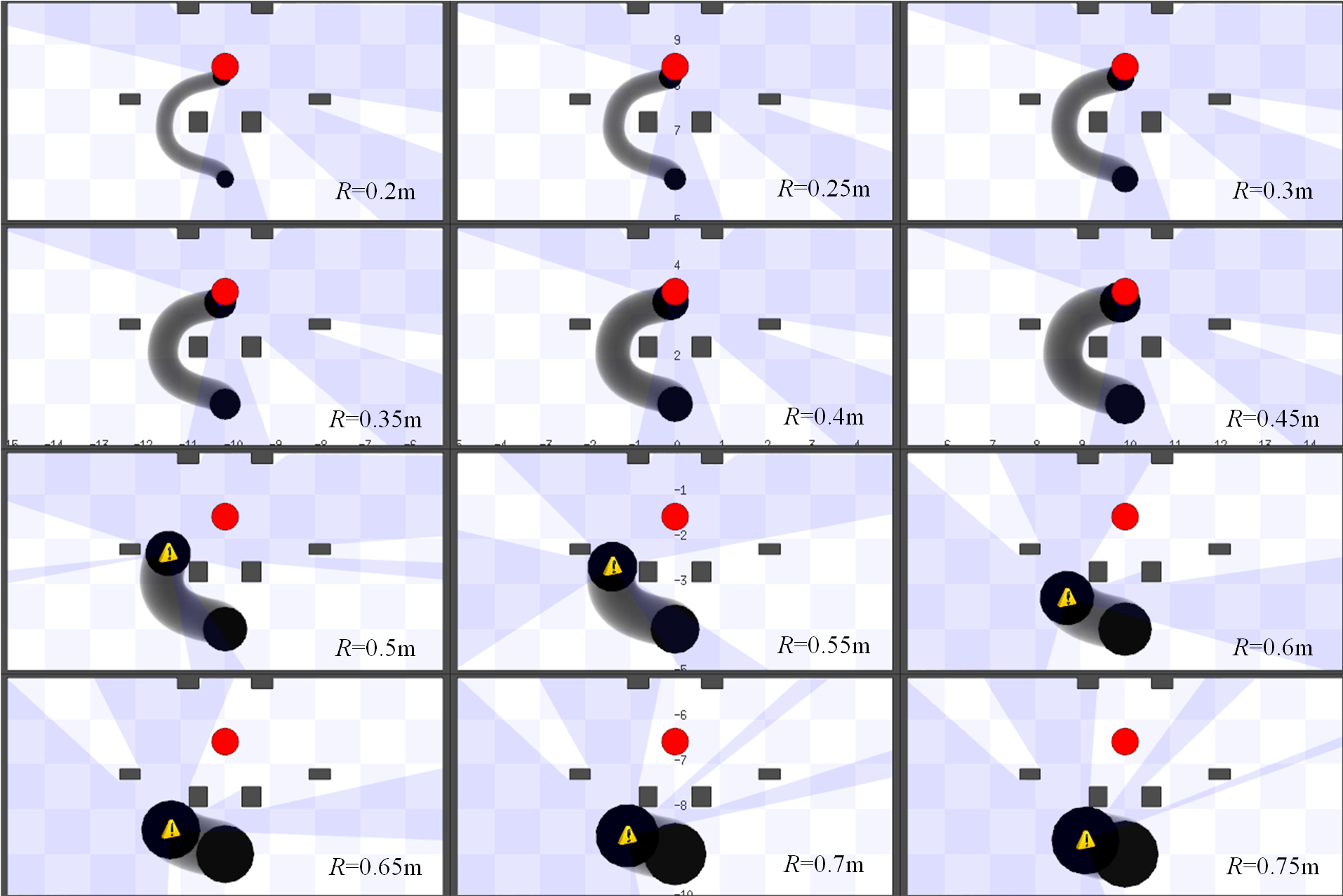}
	\caption{Trajectories of dimension-varied robots in the testing scenario, utilizing SAC\_RI as the controller.}
	\label{sim_test2}
\end{figure}

Compared to MSL\_DVST, for circular robots, an alternative approach to tackle the robot dimension-variable problem is to use SAC to train a robot with its radius as an additional input, which was referred to as SAC\_RI in this paper. We implemented SAC\_RI by training a SAC agent in a room with the same layout as the room in Fig. \ref{train1} but with a larger size ($12\times12\text{m}^2$). The DNN structure used in SAC\_RI was identical to that of MSL, with the exception that the robot's radius was included as part of the input. At the start of each episode, the robot radius is randomly sampled from the uniform distribution\textit{  \( U\left[ 0.2m,0.5m \right]  \) }. The rest training procedure is the same as MSL without curriculum learning.  Subsequently, the trained agent underwent testing to perform the same tasks depicted in Fig. \ref{sim_test}. The resulting trajectories of the robot were plotted in Fig. \ref{sim_test2}. As shown, when  \( R<0.3m \), the robot was capable of accomplishing the tasks; however, the paths taken were suboptimal. Moreover, when \( R \)  exceeded 0.5m, the robot consistently collided with obstacles, indicating that this approach was unsuitable for robots whose sizes lay outside the training range. The comparative analysis revealed the following findings:\par

\begin{enumerate}
	\item When the robot radius was small, MSL\_DVST exhibited the ability to generate more optimal paths compared to SAC\_RI.\par
	
	\item Compared to the SAC\_RI method, the MSL\_DVST method showed superior generalization ability in adapting to robots with larger dimensions.\par
\end{enumerate}\par

\section{Real-world Performance Evaluation}\label{real-experiment}
To further investigate the navigation capability of the MSL\_DVST controller, a series of real-world experiments were conducted. Notably, the controller was exclusively trained within a simulation environment, while the real-world test scenarios remained unknown to the agent.

\subsection{Mobile Robots}
\begin{figure}[t]
    \centering
	  \subfloat[]{
        \includegraphics[width=0.18\linewidth]{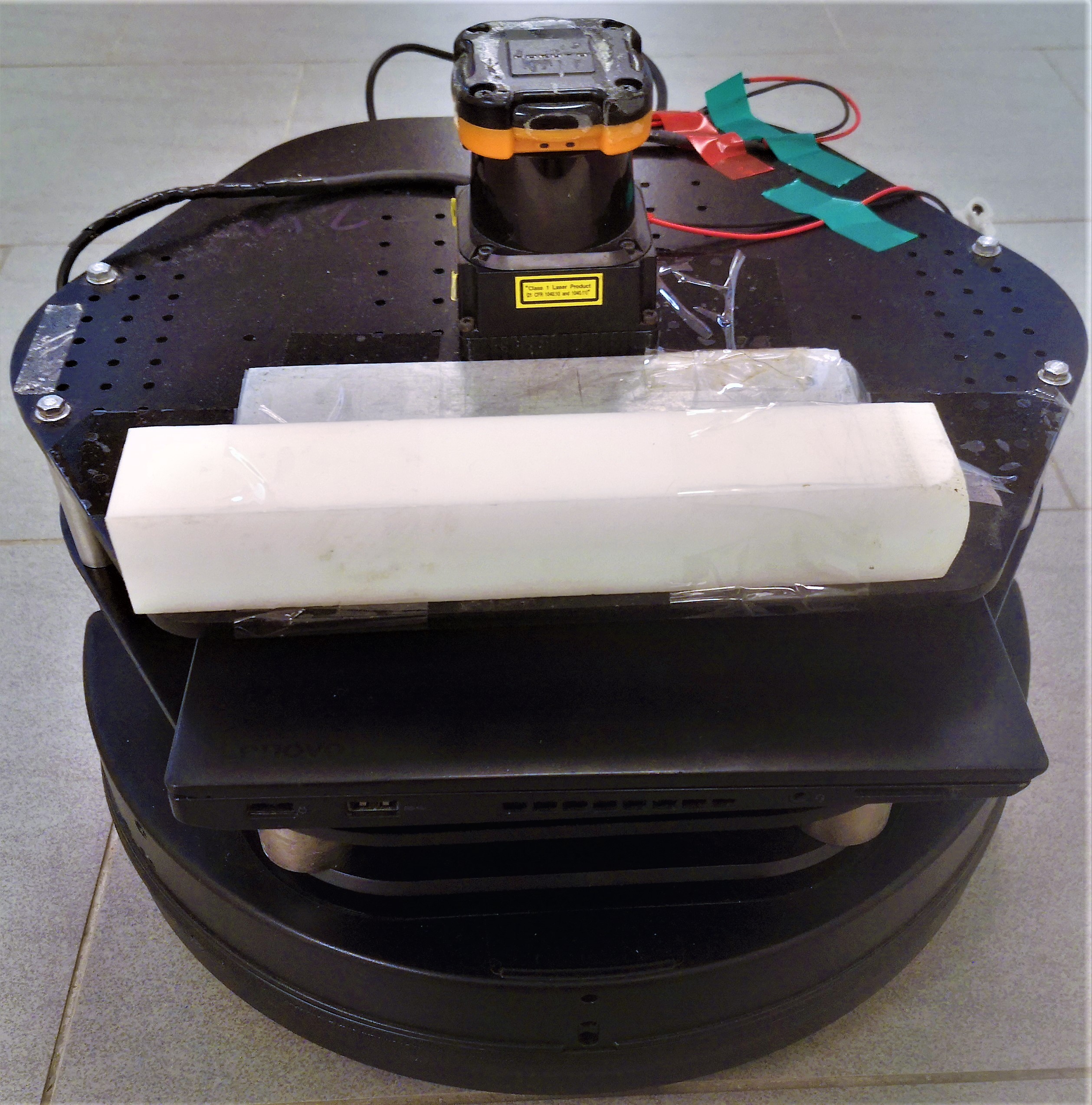}\label{r1}}
	  \subfloat[]{
        \includegraphics[width=0.33\linewidth]{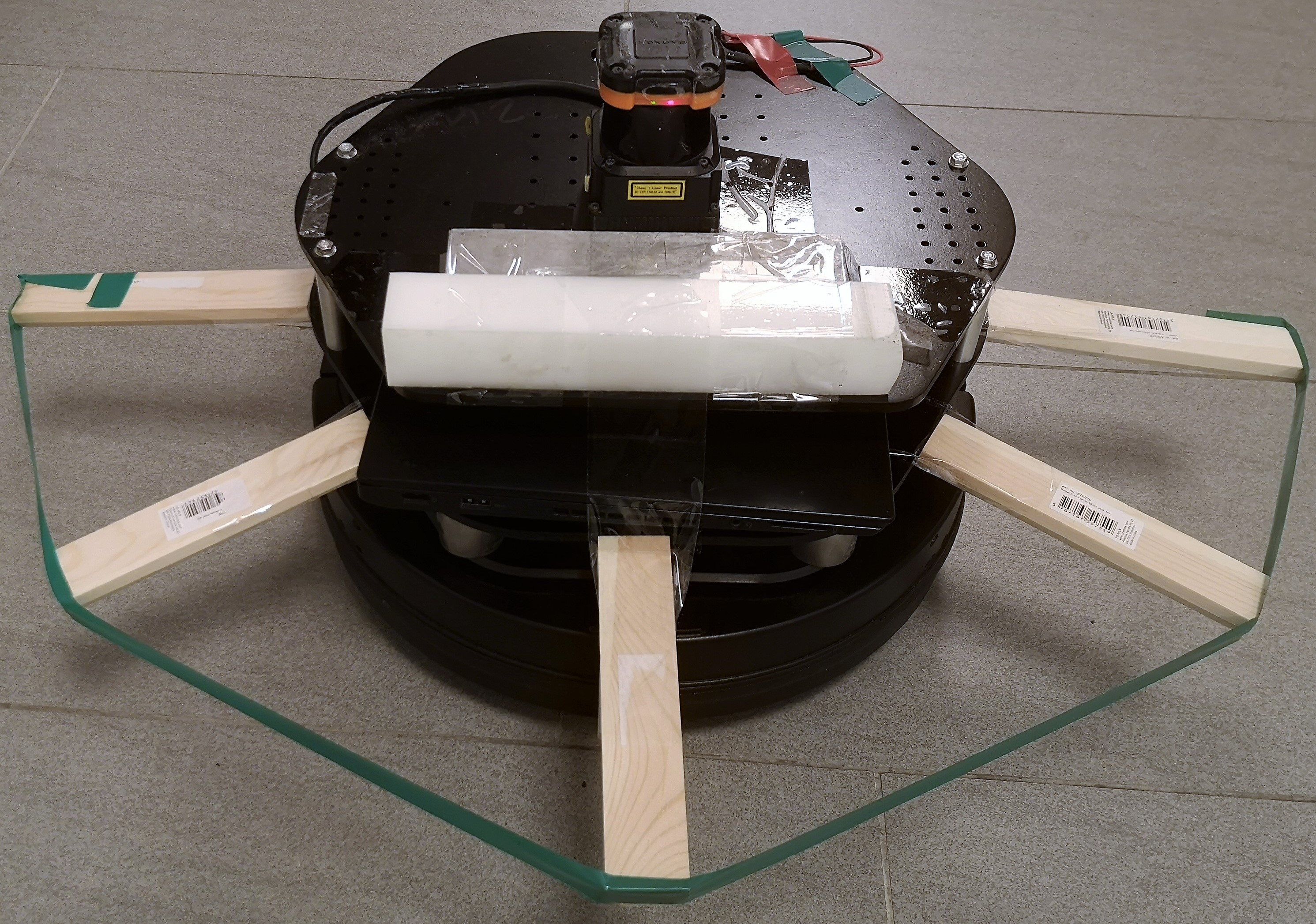}\label{r2}}
        \subfloat[]{
        \includegraphics[width=0.46\linewidth]{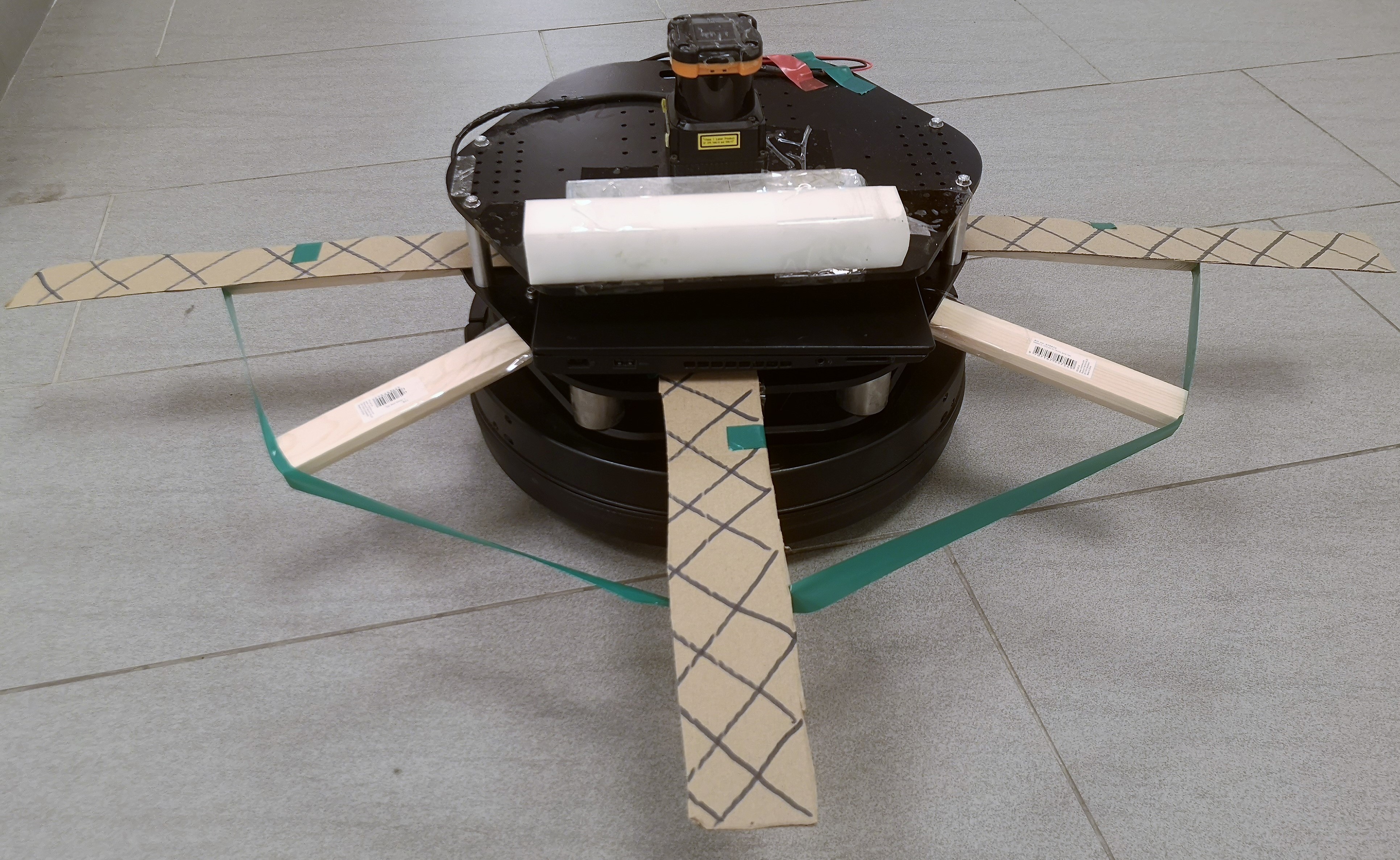}\label{r3}}
	\caption{The robots used in real-world testing: (a) the small robot, $R$=$0.2$m; (b) the medium-sized robot, $R$=$0.37$m; (c) the large robot, $R$=$0.52$m.}
	\label{robt_tested}
\end{figure}

Turtlebot2 was employed as the mobile robot for conducting the real-world experiment. As shown in Fig. \ref{r1}, it was equipped with a Hokuyo UTM-30LX Lidar and controlled by a laptop. Its radius was $17.7\text{cm}$ (small robot). Besides, we built two dimension-varied robots on turtlebot2, namely the medium-sized robot and the large robot, as shown in Fig. \ref{r2} and \ref{r3}. To apply DVST, the dimension-scaled robot corresponding to the medium-sized robot had a radius of $34\text{cm}$, while the dimension-scaled robot corresponding to the large robot had a radius of $49\text{cm}$. Considering the sim-to-real gap such as sensor noise, actuation delays, and dynamic constraints, the one-step trajectory generated by the real robot could not be as perfect as its simulated counterpart. Fortunately, the trajectory error would not accumulate because the navigation agent would adjust its policy based on newly obtained observations at each decision step. To mitigate potential collisions resulting from the one-step trajectory error, we augmented the robot radii when employing DVST. Specifically, the safety radii employed for the small, medium-sized, and large robots were $20\text{cm}$, $37\text{cm}$, and $52\text{cm}$, respectively. Due to the absence of sensors that could directly obtain the target position, the same as the approach employed in \cite{xie2021} and \cite{zhang2022ipaprec}, the AMCL ROS package (AMCL) \cite{AMCL} was employed to localize the robot within a pre-built global map. By leveraging the robot's positions derived from the map, the relative position of the goal in the robot's frame could be determined. It is noteworthy that the map itself was not provided to the navigation agent.

\subsection{Testing in Real-world Static Scenarios}

\begin{figure}[t]
    \centering
	  \subfloat[]{
        \includegraphics[width=0.48\linewidth]{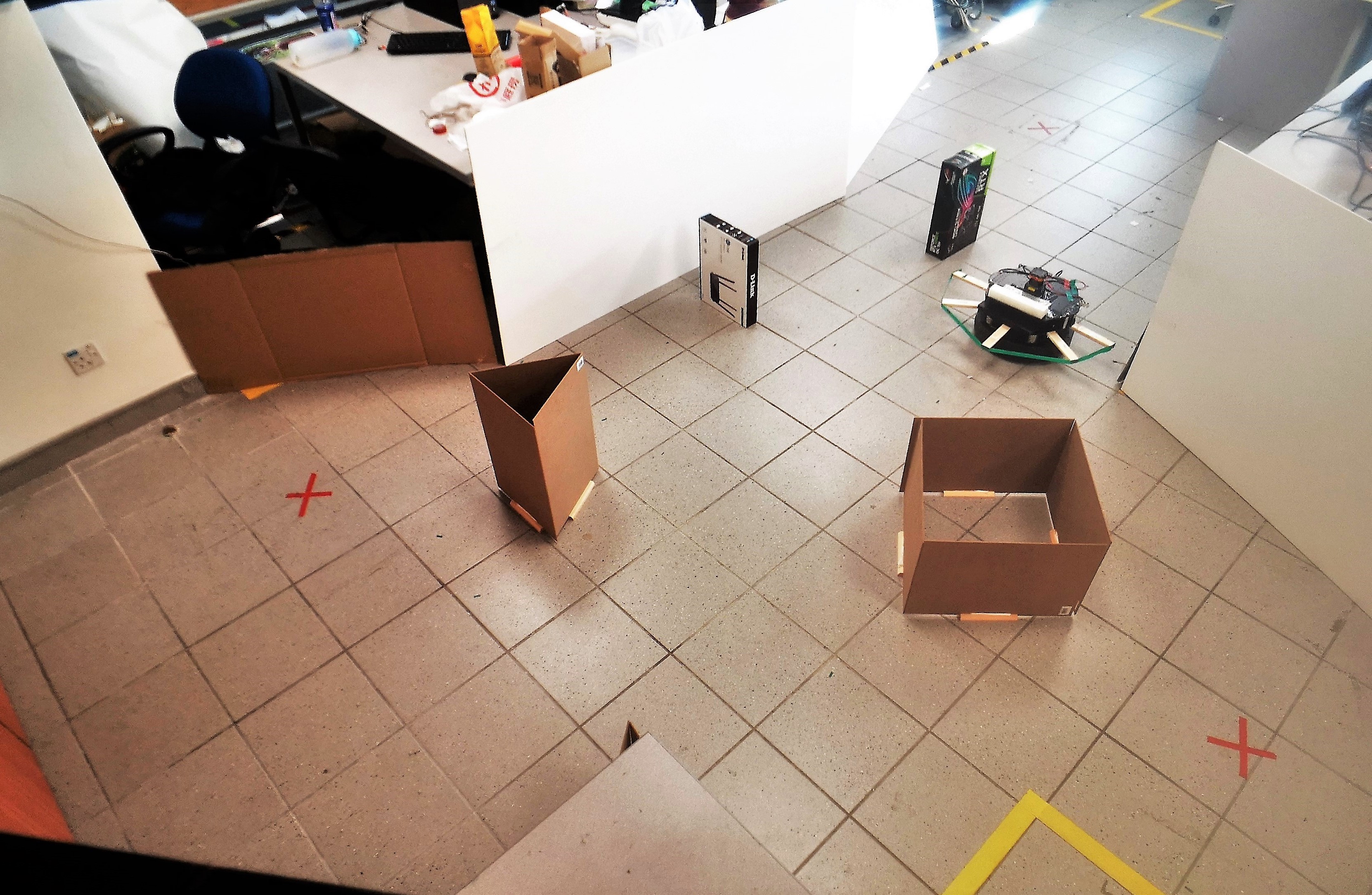}\label{env0}}
	  \subfloat[]{
        \includegraphics[width=0.48\linewidth]{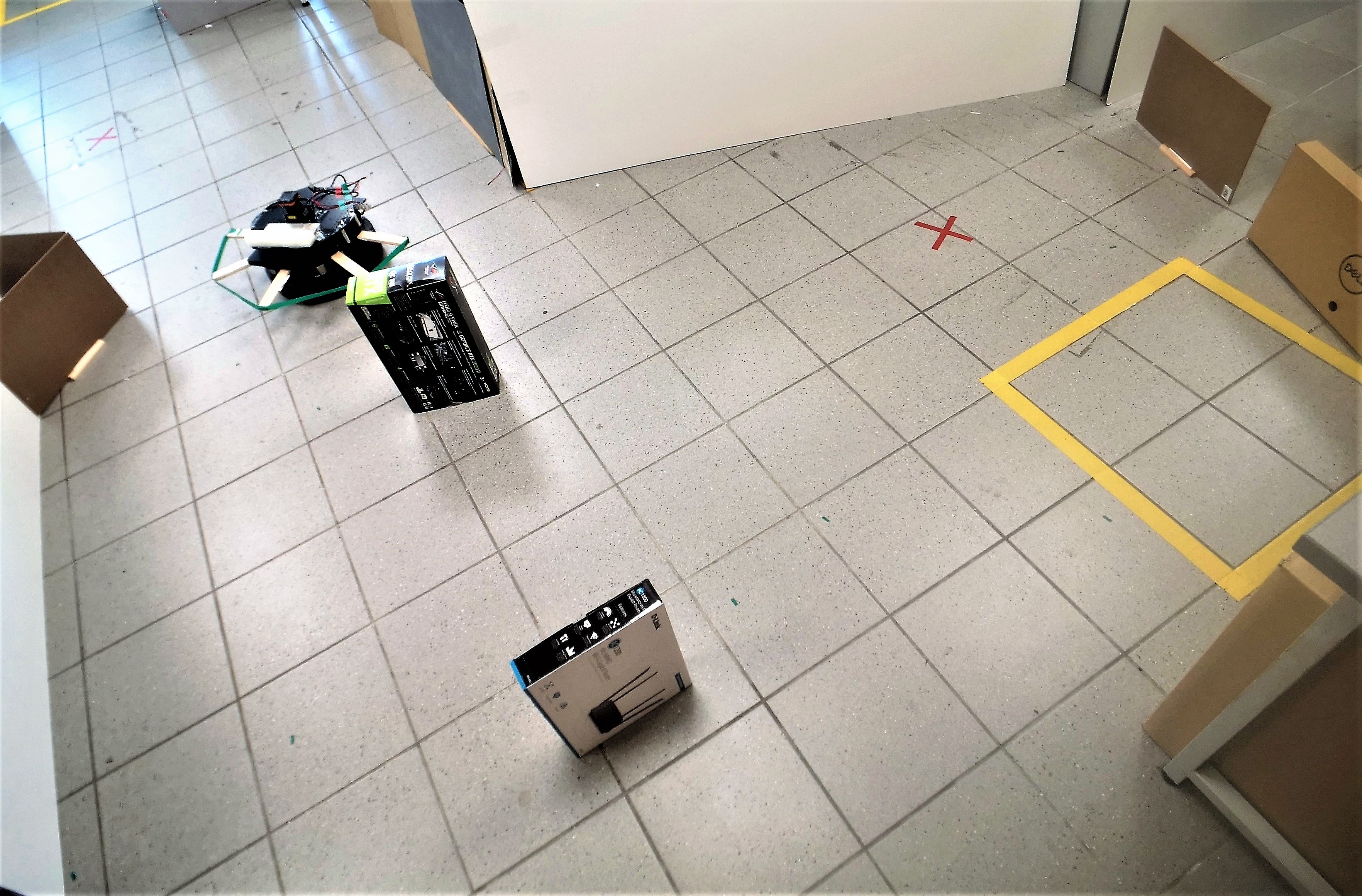}\label{env1}}
        \newline
        \subfloat[]{
        \includegraphics[width=0.48\linewidth]{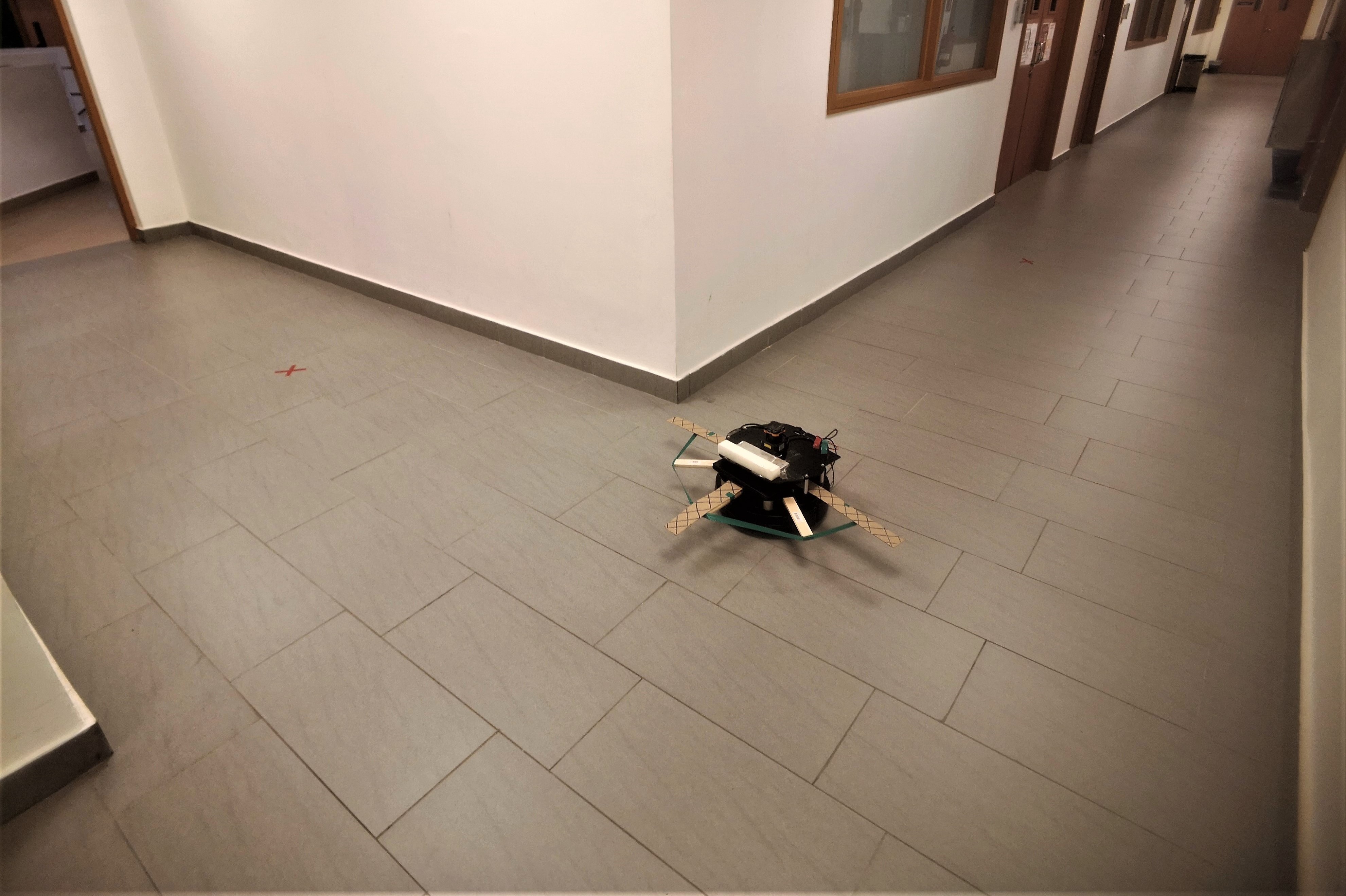}\label{env2}}
        \subfloat[]{
        \includegraphics[width=0.48\linewidth]{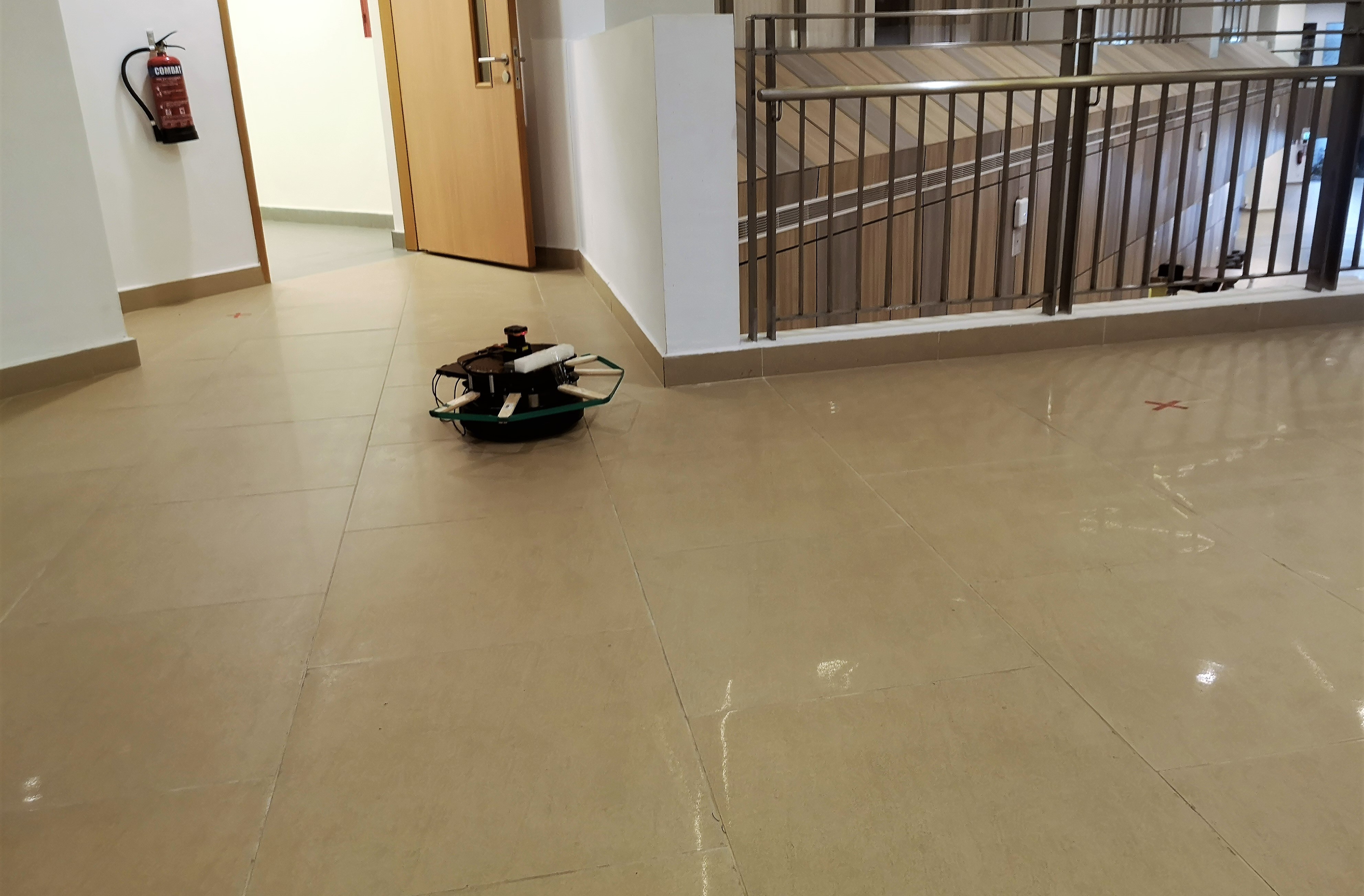}\label{env3}}
	\caption{The real-world testing scenarios with static obstacles: (a) REnv\_0; (b) REnv\_1; (c) REnv\_2; (d) REnv\_3.}
	\label{real_env}
\end{figure}

As shown in Fig. \ref{real_env}, four real-world scenarios were used to test the navigation performance of robots. In two indoor scenarios (REnv\_0 and REnv\_1), the obstacles were meticulously placed to ensure that there existed a path allowing the medium-sized robot to pass. In each scenario, the goal points were marked by red crosses. Due to spatial constraints, the large robot was only tested in the corridor scenario (REnv\_2) displayed in Fig. \ref{env2}. In each task, the primary objective of the robot was to sequentially navigate towards multiple designated goal points.

\begin{figure}[t]
    \centering
	  \subfloat[]{
        \includegraphics[width=0.48\linewidth]{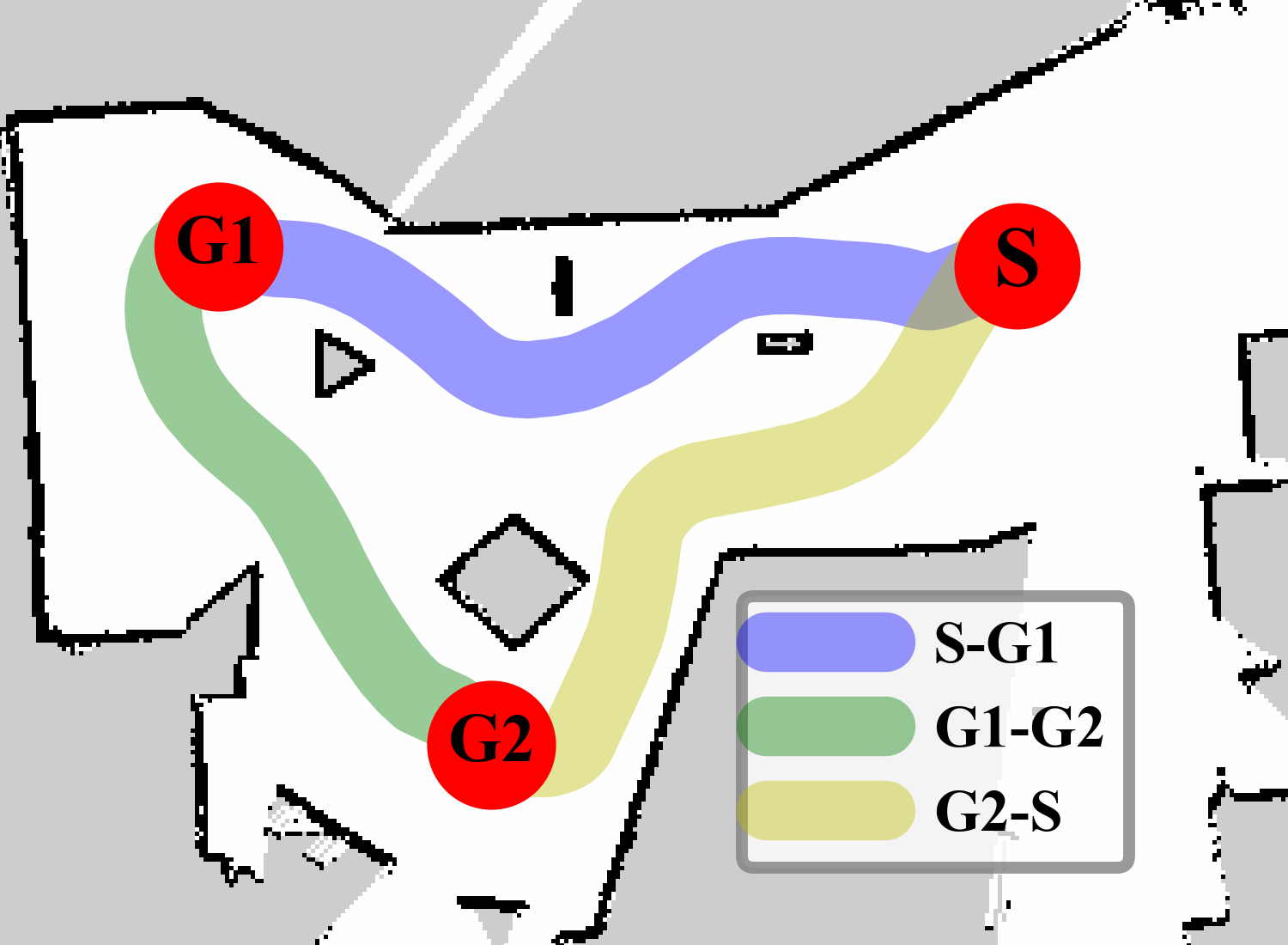}\label{indoor0}}
	  \subfloat[]{
        \includegraphics[width=0.48\linewidth]{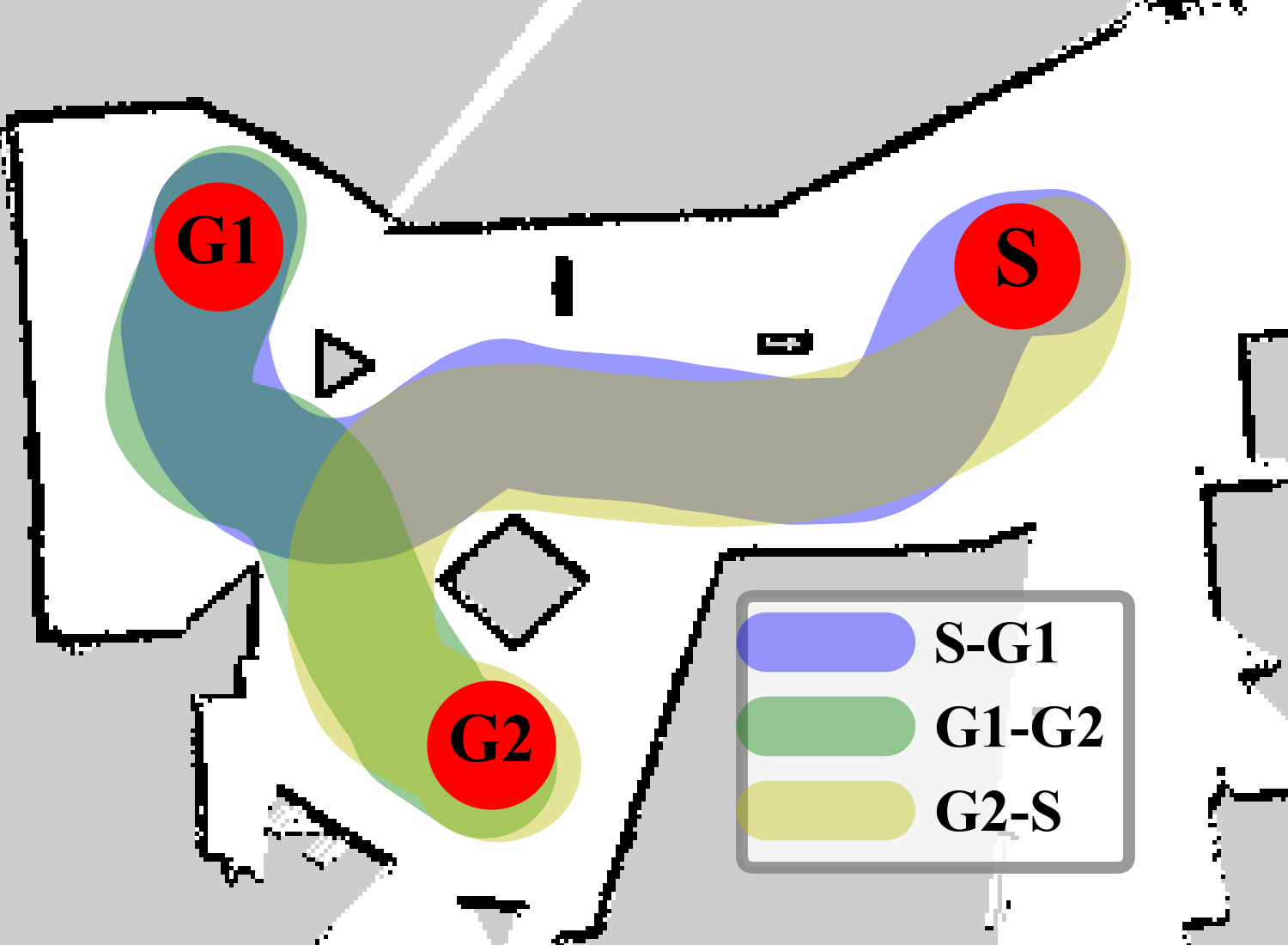}\label{indoor2}}
        \newline
        \subfloat[]{
        \includegraphics[width=0.48\linewidth]{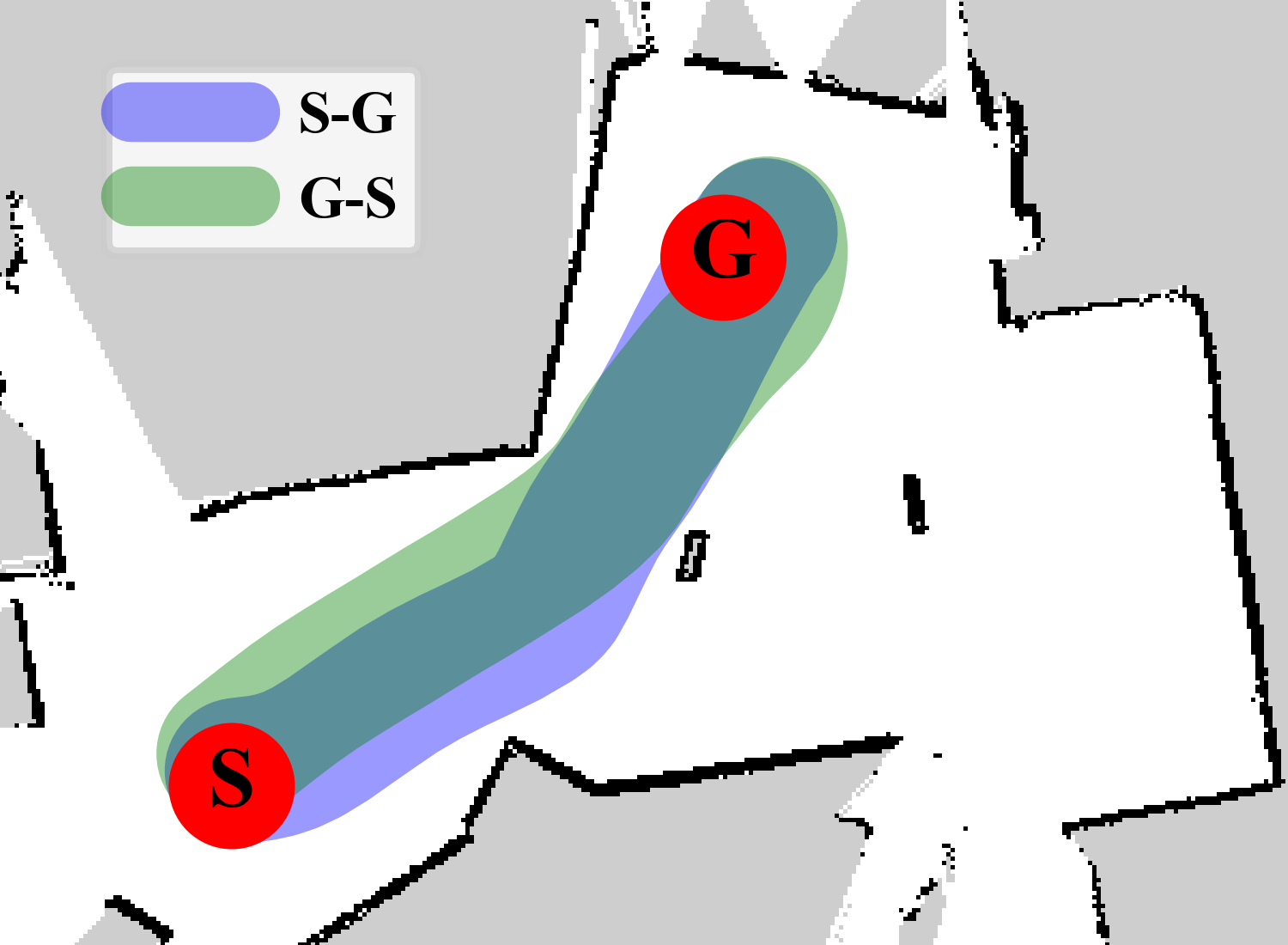}\label{dynamic0}}
        \subfloat[]{
        \includegraphics[width=0.48\linewidth]{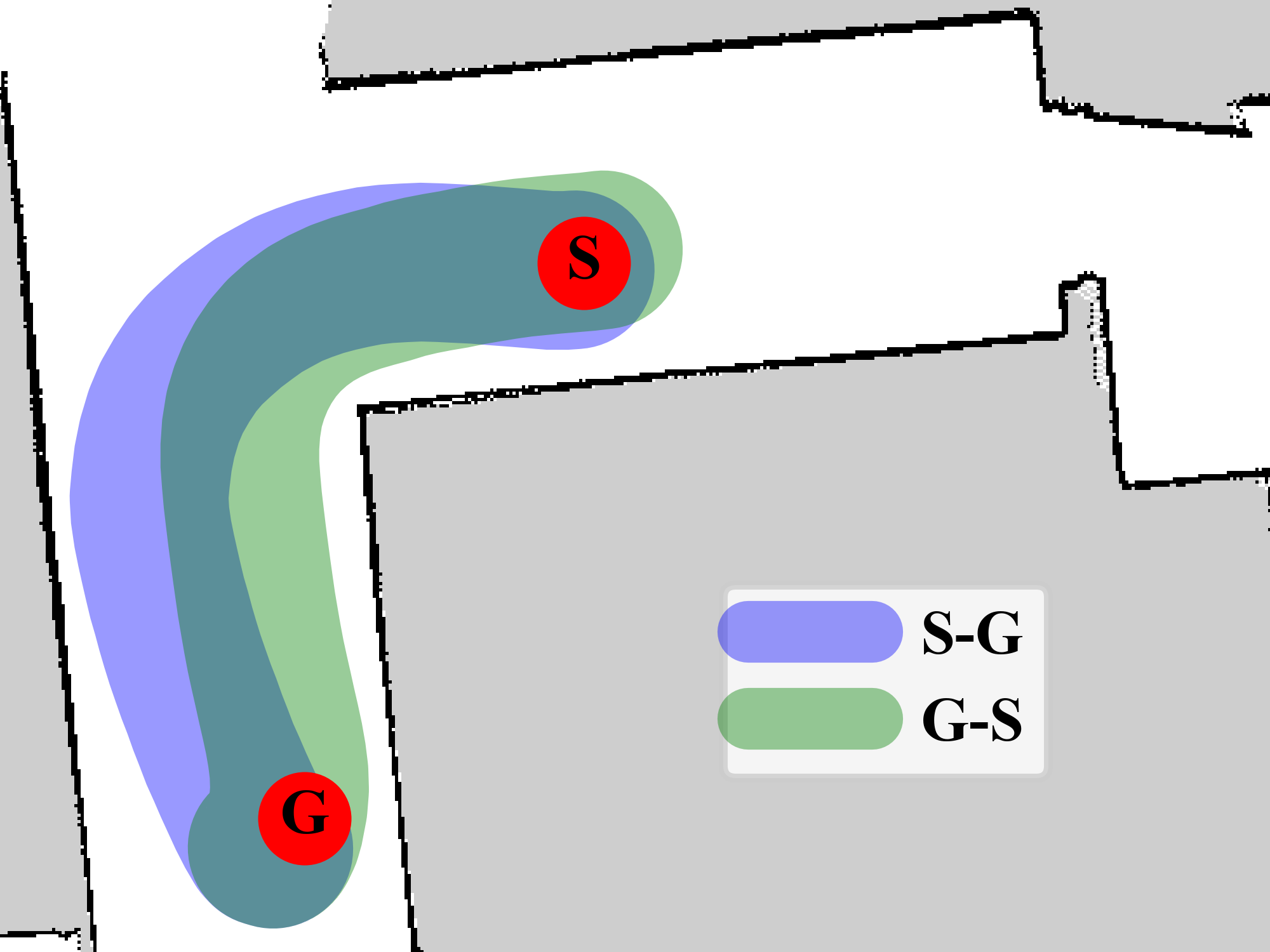}\label{out2}}
        \newline
        \subfloat[]{
        \includegraphics[width=0.48\linewidth]{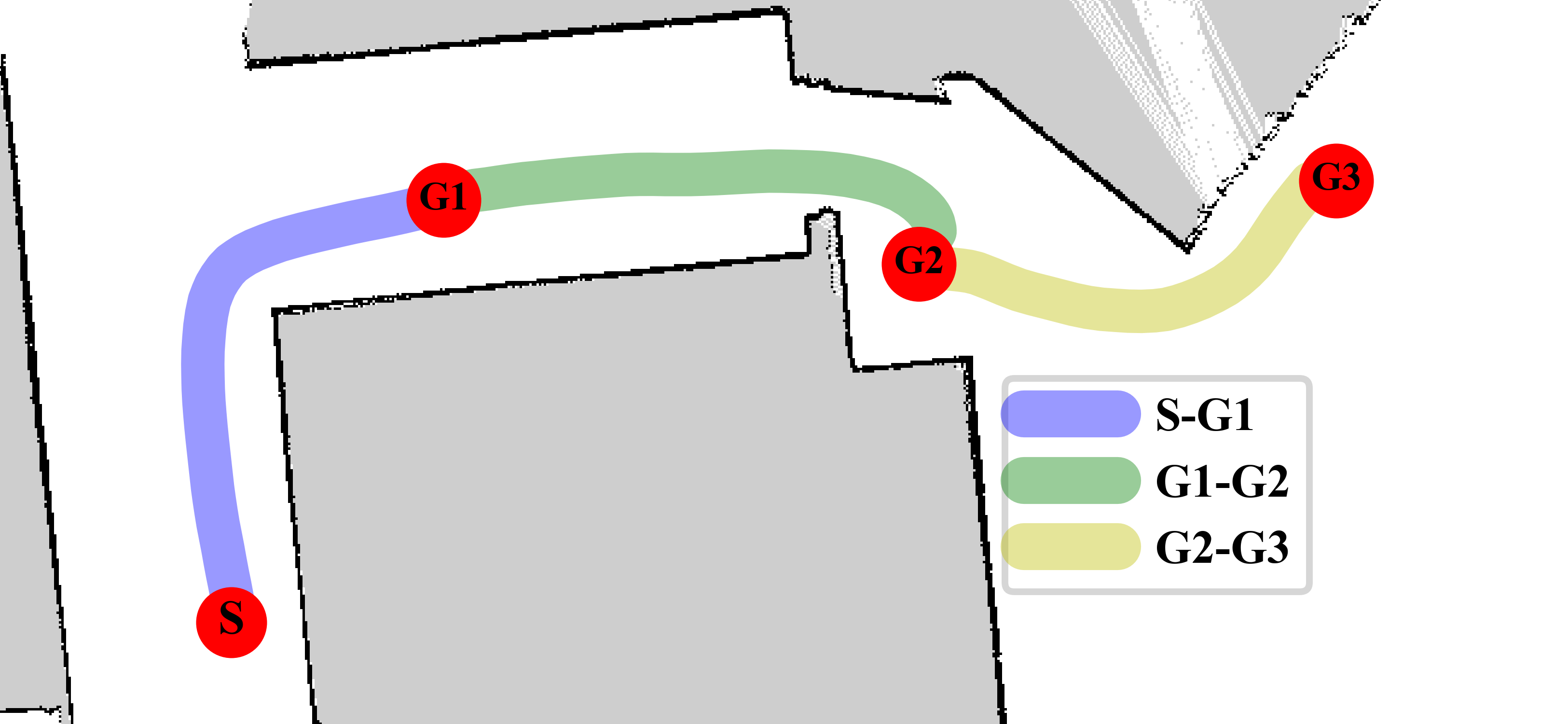}\label{out0}}
        \subfloat[]{
        \includegraphics[width=0.48\linewidth]{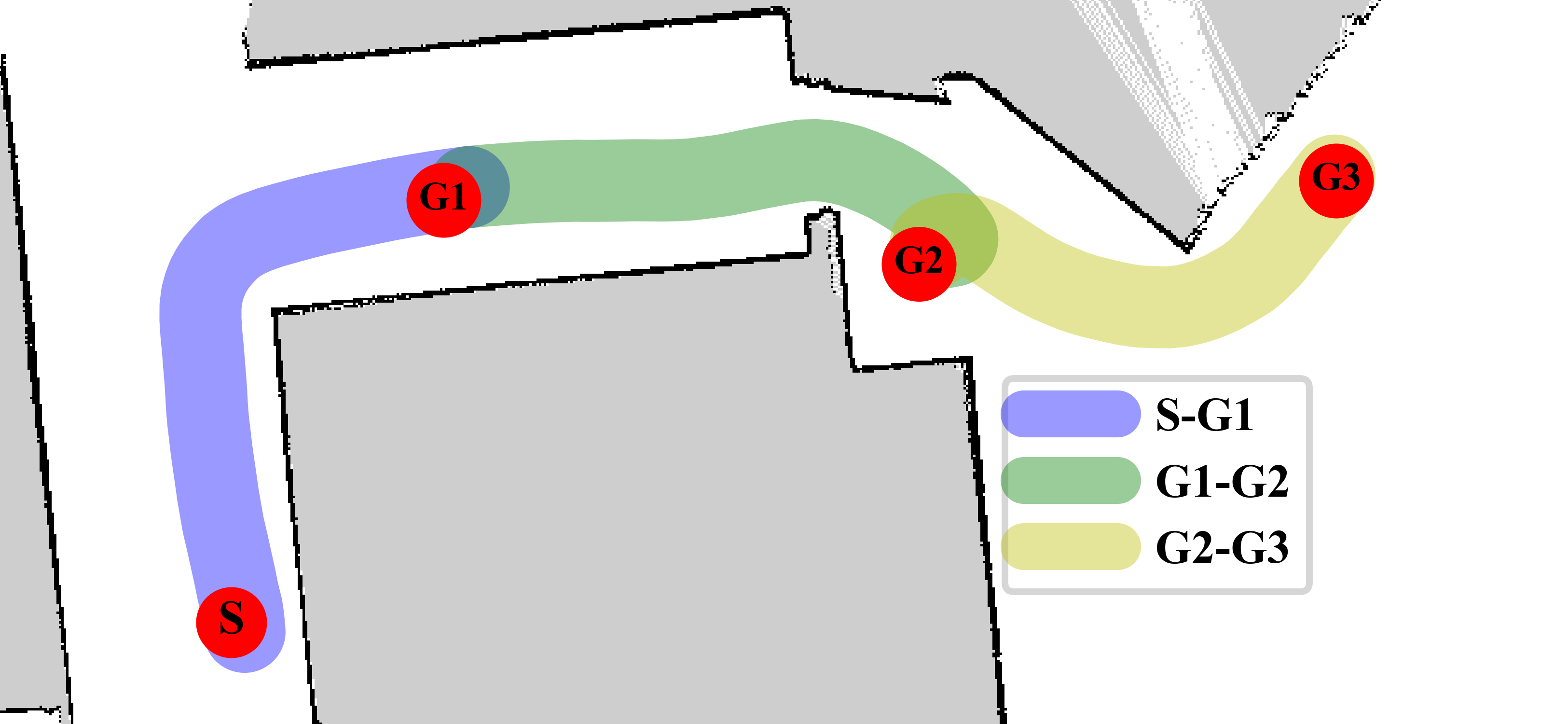}\label{out1}}
	\caption{The trajectories of the tested robots in different testing scenarios. The task descriptions for sub-figures were listed in Table \ref{task_list}.}
	\label{tested_result}
\end{figure}

The scenario setting and completion time for each task were indicated in Table \ref{task_list}. The navigation videos for these tasks can be found in the supplementary file, while the resulting robot trajectories were plotted in Fig. \ref{tested_result}. As shown, all six tasks were successfully accomplished. When the robot radius was small ($R$=$0.2$m, see Fig. \ref{indoor0}), the robot  selected shorter paths to reach the target. When the robot radius increased to $R$=$0.37$m, as shown in Fig. \ref{indoor2}, the robot recognized its inability to traverse the narrow passages and consequently transitioned to wider ones. As shown in Table \ref{task_list}, when in relative open scenarios, such as Task (e) and (f), the medium-sized robot achieved comparable performance to the small robot in terms of navigation time. It indicated that the dimension-varied robot could perform navigation tasks at a performance level comparable to the meta robot\par

\begin{table}[]
\caption{\label{task_list} Real-world navigation tasks and completion time $T_{task}$.} 
\renewcommand\arraystretch{1.2}
\centering
% Please add the following required packages to your document preamble:
% \usepackage{multirow}
\setlength{\tabcolsep}{1.4mm}{\begin{tabular}{lllll}
\hline
\hline
Task & Testing scenario & Robot radius  & $T_{task}$ \\ \hline
(a) & REnv\_0 & 0.2m  & 32s \\
(b) & REnv\_0 & 0.37m  & 47s \\
(c) & REnv\_1 & 0.37m  & 21s \\
(d) & REnv\_2 & 0.52m  & 28s \\
(e) & REnv\_2\&3 & 0.2m  & 40s \\
(f) & REnv\_2\&3 & 0.37m  & 37s \\ \hline \hline
\end{tabular}}
\end{table}

\subsection{Reaction to Sudden Changes}
\begin{figure}[t]
    \centering
	  \subfloat[]{
        \includegraphics[width=0.48\linewidth]{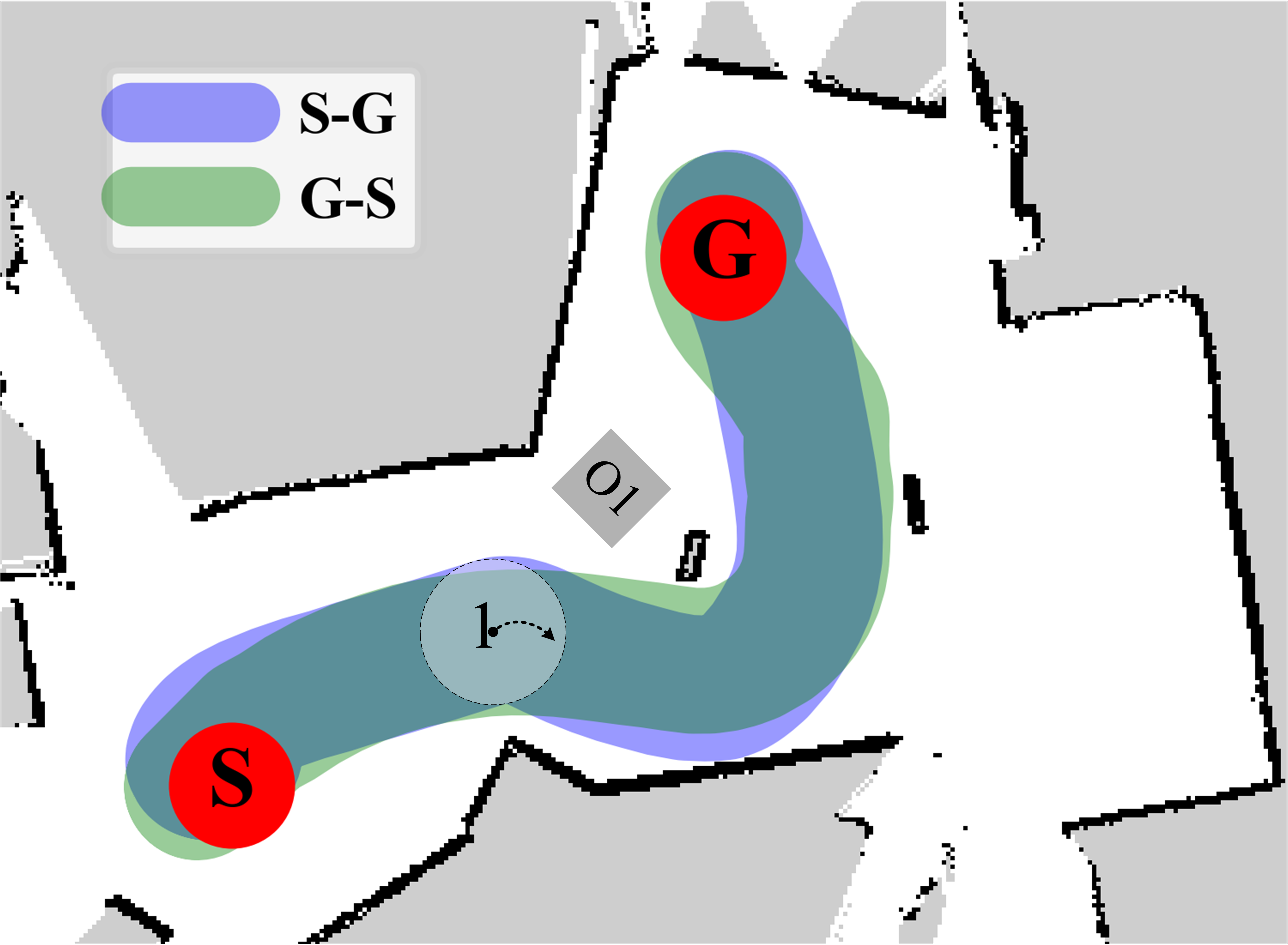}\label{dynamic1}}
	  \subfloat[]{
        \includegraphics[width=0.48\linewidth]{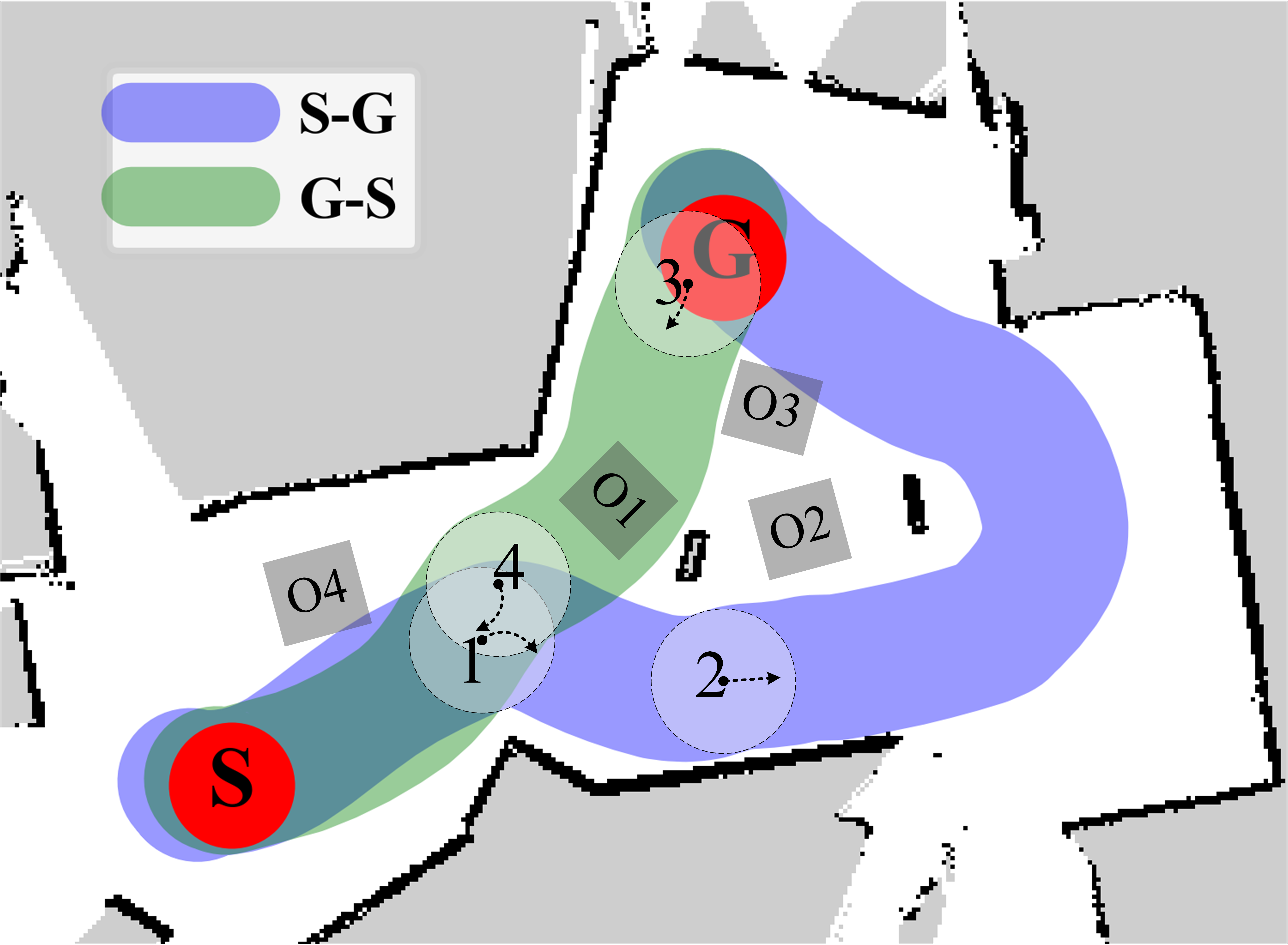}\label{dynamic2}}
	\caption{The trajectories of medium-sized robot ($R$=$0.37$m) in REnv\_1 when encountering obstacles that suddenly pop up.}
	\label{dynamic_test}
\end{figure}

Finally, the performance of the MSL\_DVST method was assessed when the robot encountered suddenly appearing obstacles (the tester and a wooden obstacle), as depicted in Fig. \ref{dynamic_env}. The full video of the experiments can be found in the supplementary file, while the resulting trajectories of the medium-sized robot tested in REnv\_1 were plotted in Fig. \ref{dynamic_test}. For a better description, when the obstacle $``$O$i$$"$ suddenly appeared, we marked the robot position at that time as point $``$$i$$"$. As shown in Fig. \ref{dynamic1}, when the robot moved to point $``$$1$$"$, its optimal path was plotted in Fig. \ref{dynamic0}. However, the tester abruptly blocked the optimal path with obstacle $``$O$1$$"$, and the robot changed its direction by turning right and bypassed the obstacles.  Moreover, in Fig. \ref{dynamic2}, the tester successively blocked two paths using obstacles $``$O$1$$"$ and $``$O$2$$"$. Although facing such abrupt changes, the robot adaptively chose the only path and reached the goal $``$G$"$. The results obtained in such dynamic scenarios indicated that the dimension-varied robots could well handle sudden changes during navigation.
\begin{figure}[t]
    \centering
	  \subfloat[]{
        \includegraphics[width=0.48\linewidth]{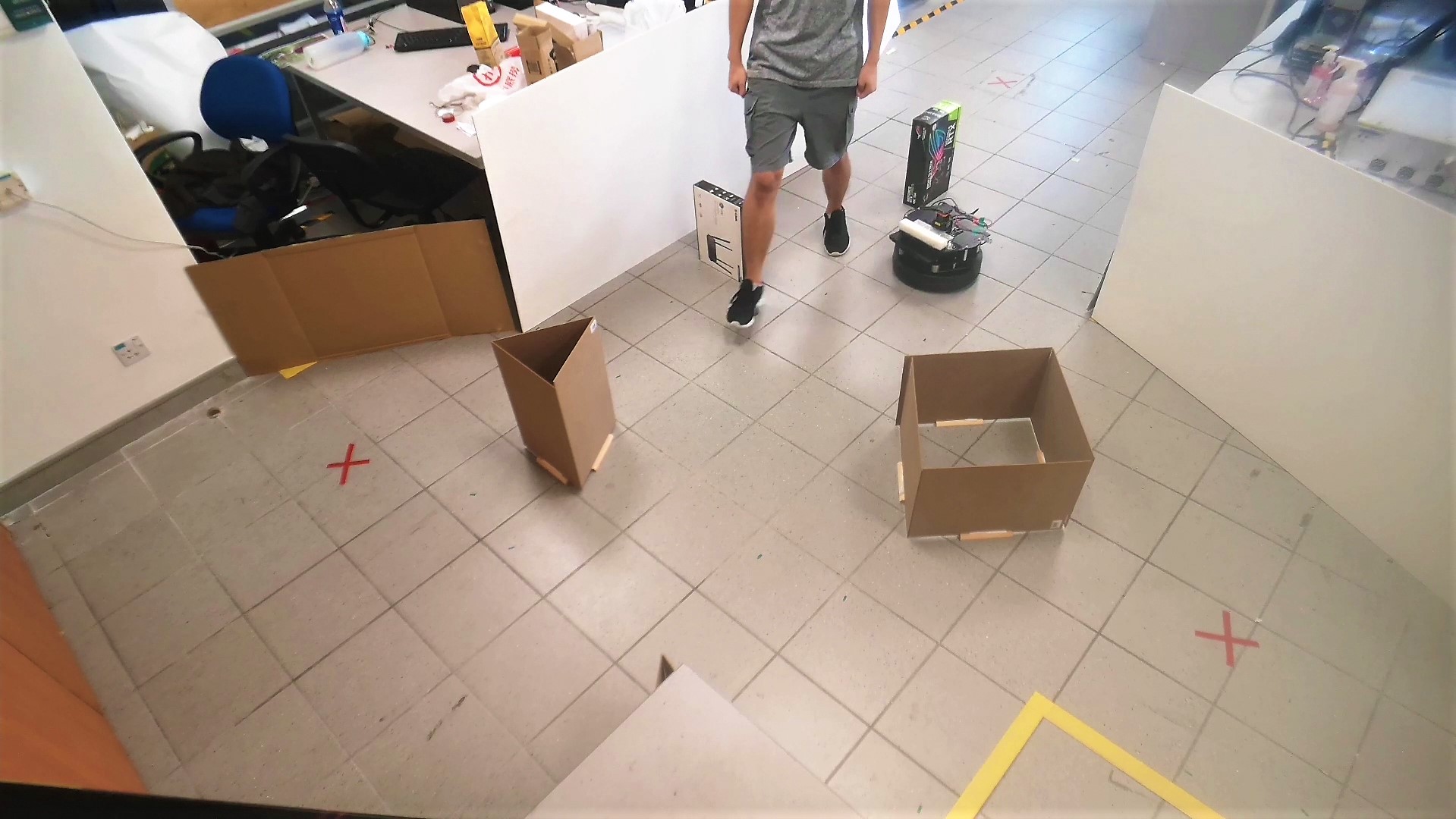}\label{d1}}
	  \subfloat[]{
        \includegraphics[width=0.48\linewidth]{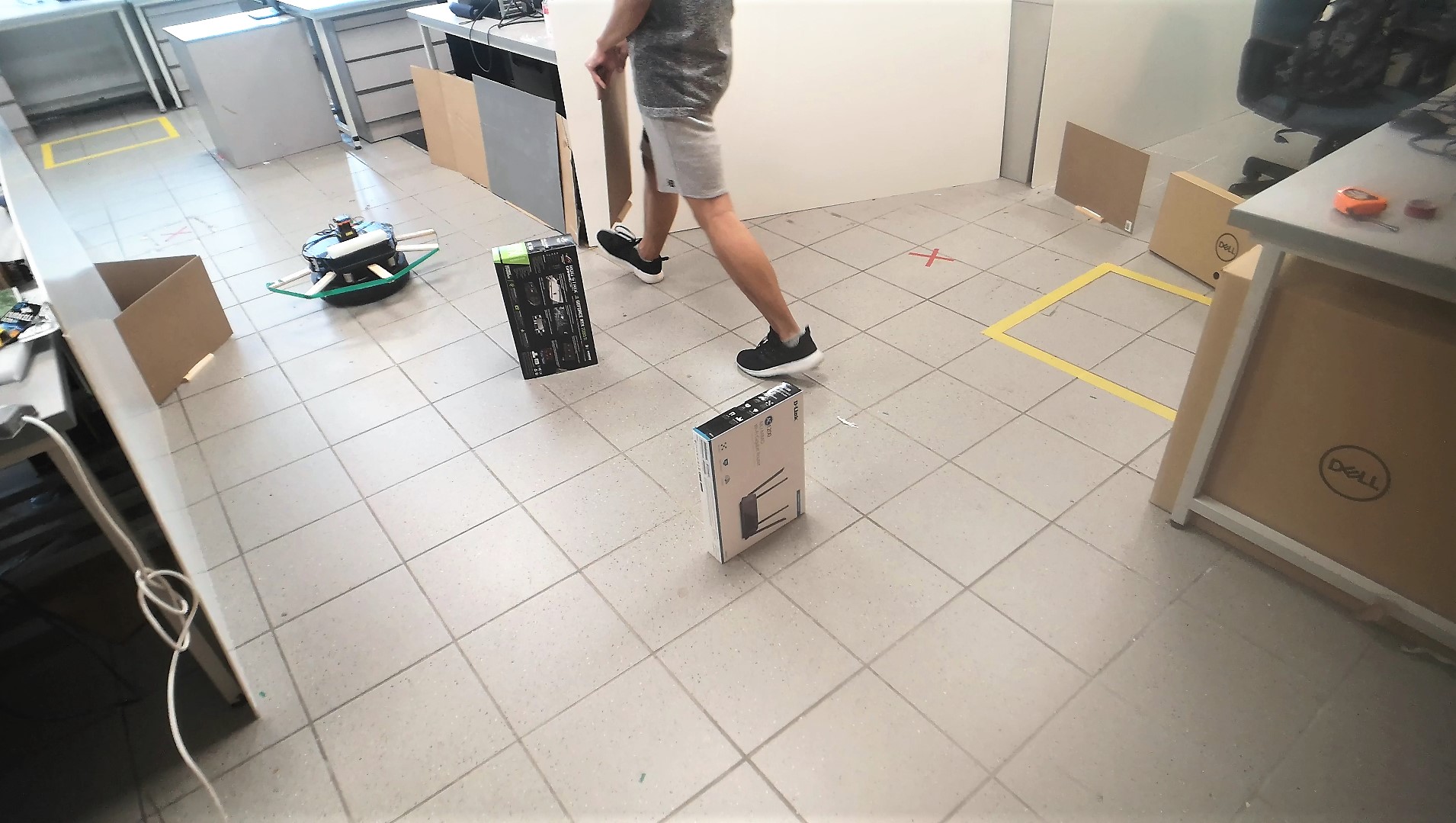}\label{d2}}
        \newline
        \subfloat[]{
        \includegraphics[width=0.48\linewidth]{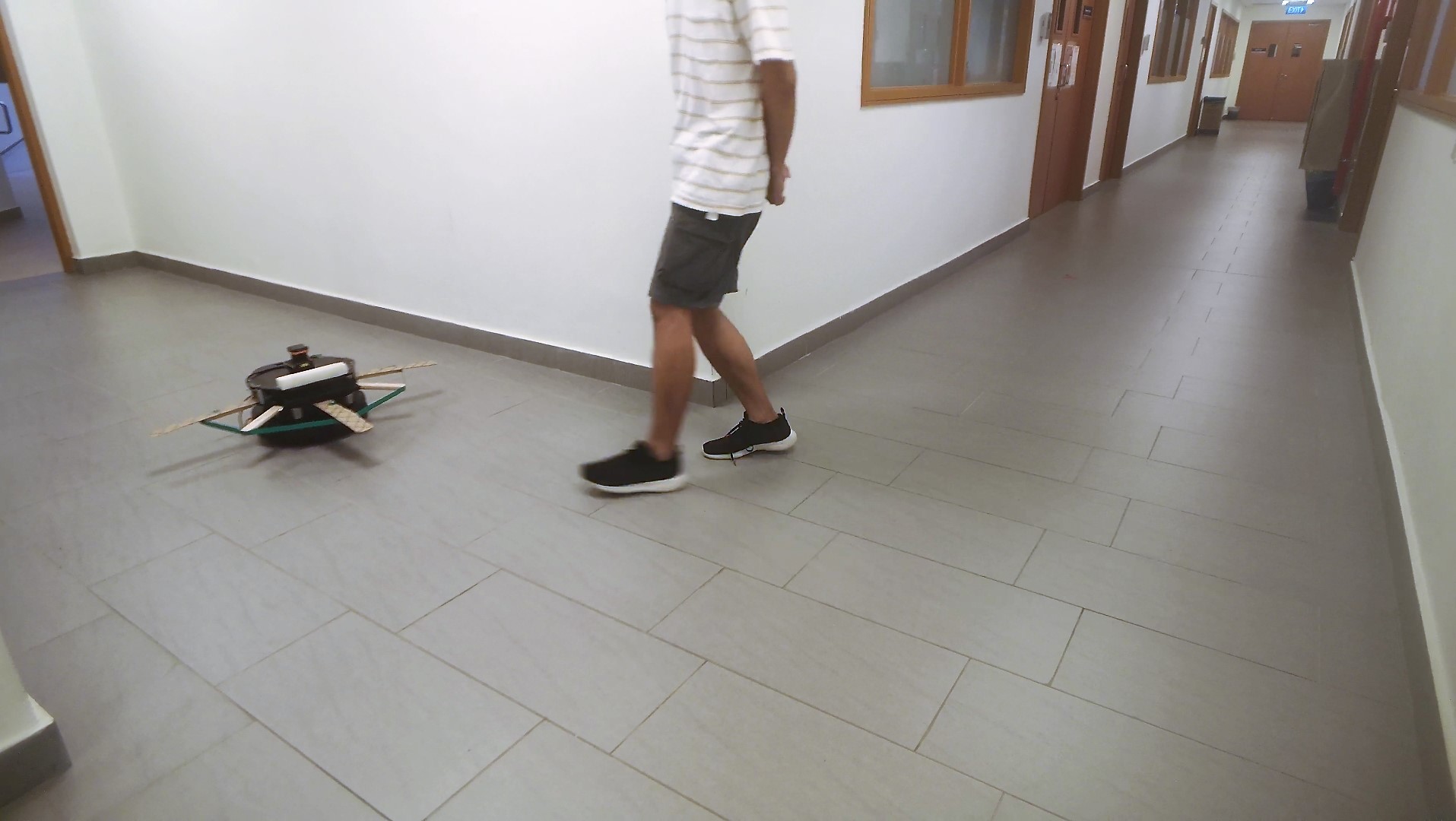}\label{d3}}
        \subfloat[]{
        \includegraphics[width=0.48\linewidth]{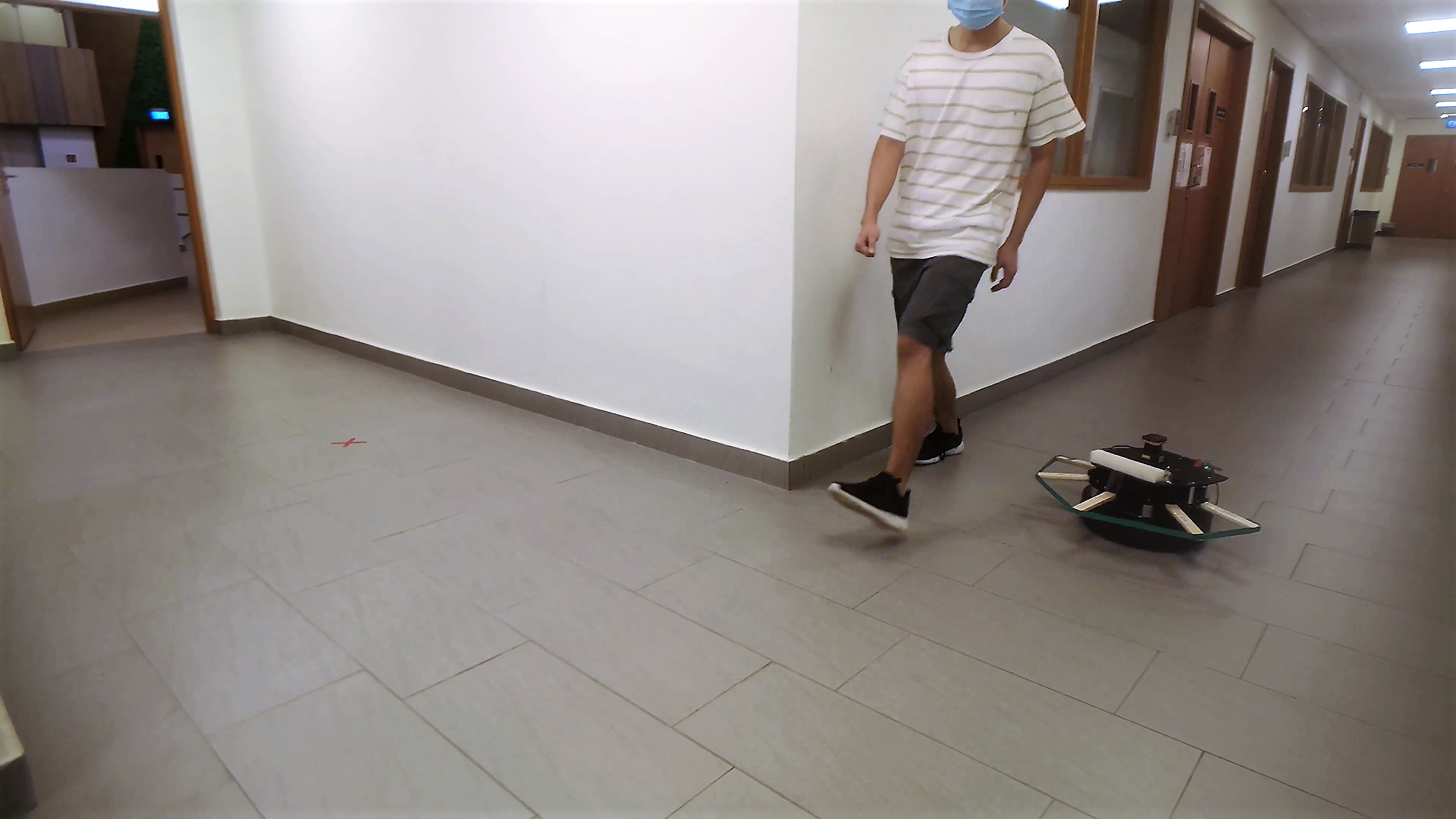}\label{d4}}
	\caption{The real-world testing scenarios with suddenly appearing obstacles.}
	\label{dynamic_env}
\end{figure}

\section{Conclusion}\label{conclusion}
In this paper, we presented MSL\_DVST, a dimension-variable DRL-based robot navigation method, specifically designed to handle scenarios where the robot needs to transport cargo or victims larger than its physical dimensions. The real-world experimental results demonstrated the feasibility of deploying the MSL\_DVST-based controllers into dimension-varied robots without any retraining. The proposed method substantially expanded the applicability of DRL-based navigation controllers from fixed dimensions to varied dimensions, thereby enhancing the robot's flexibility during task execution. Moreover, the dimension-varied robots exhibited efficient navigation capabilities in unknown and dynamic environments. The utilization of dimension-scaled robots ensures safe skill transfer; however, this approach compromises the optimality of the control policy, particularly when the shape of dimension-varied robots substantially diverges from the meta robot. To address this issue, in the future, we plan to improve the DVST method to generate more optimal policies for dimension-varied robots.\par

\ifCLASSOPTIONcaptionsoff
  \newpage
\fi

\ifCLASSOPTIONcaptionsoff
  \newpage
\fi

% trigger a \newpage just before the given reference
% number - used to balance the columns on the last page
% adjust value as needed - may need to be readjusted if
% the document is modified later
%\IEEEtriggeratref{8}
% The "triggered" command can be changed if desired:
%\IEEEtriggercmd{\enlargethispage{-5in}}

% references section

% can use a bibliography generated by BibTeX as a .bbl file
% BibTeX documentation can be easily obtained at:
% http://mirror.ctan.org/biblio/bibtex/contrib/doc/
% The IEEEtran BibTeX style support page is at:
% http://www.michaelshell.org/tex/ieeetran/bibtex/
%\bibliographystyle{IEEEtran}
% argument is your BibTeX string definitions and bibliography database(s)
%\bibliography{IEEEabrv,../bib/paper}
%
% <OR> manually copy in the resultant .bbl file
% set second argument of \begin to the number of references
% (used to reserve space for the reference number labels box)

\bibliographystyle{IEEEtran}
\bibliography{IEEEabrv,BIB_xx-ZW-xxxx}\

% Generated by IEEEtran.bst, version: 1.14 (2015/08/26)
\begin{thebibliography}{10}
\providecommand{\url}[1]{#1}
\csname url@samestyle\endcsname
\providecommand{\newblock}{\relax}
\providecommand{\bibinfo}[2]{#2}
\providecommand{\BIBentrySTDinterwordspacing}{\spaceskip=0pt\relax}
\providecommand{\BIBentryALTinterwordstretchfactor}{4}
\providecommand{\BIBentryALTinterwordspacing}{\spaceskip=\fontdimen2\font plus
\BIBentryALTinterwordstretchfactor\fontdimen3\font minus
  \fontdimen4\font\relax}
\providecommand{\BIBforeignlanguage}[2]{{%
\expandafter\ifx\csname l@#1\endcsname\relax
\typeout{** WARNING: IEEEtran.bst: No hyphenation pattern has been}%
\typeout{** loaded for the language `#1'. Using the pattern for}%
\typeout{** the default language instead.}%
\else
\language=\csname l@#1\endcsname
\fi
#2}}
\providecommand{\BIBdecl}{\relax}
\BIBdecl

\bibitem{Yue2021col}
Y.~Yue, C.~Zhao, Z.~Wu, C.~Yang, Y.~Wang, and D.~Wang, ``Collaborative semantic
  understanding and mapping framework for autonomous systems,'' \emph{IEEE/ASME
  Trans. Mechatronics}, vol.~26, no.~2, pp. 978--989, 2021.

\bibitem{Zou2022A}
Q.~Zou, Q.~Sun, L.~Chen, B.~Nie, and Q.~Li, ``A comparative analysis of lidar
  slam-based indoor navigation for autonomous vehicles,'' \emph{IEEE Trans.
  Intell. Transp. Syst.}, vol.~23, no.~7, pp. 6907--6921, 2022.

\bibitem{mnih2015human}
V.~Mnih, K.~Kavukcuoglu, D.~Silver, A.~A. Rusu, J.~Veness, M.~G. Bellemare,
  A.~Graves, M.~Riedmiller, A.~K. Fidjeland, G.~Ostrovski \emph{et~al.},
  ``Human-level control through deep reinforcement learning,'' \emph{nature},
  vol. 518, no. 7540, pp. 529--533, 2015.

\bibitem{Wang2023}
H.-C. Wang, S.-C. Huang, P.-J. Huang, K.-L. Wang, Y.-C. Teng, Y.-T. Ko,
  D.~Jeon, and I.-C. Wu, ``Curriculum reinforcement learning from avoiding
  collisions to navigating among movable obstacles in diverse environments,''
  \emph{IEEE Robot. Autom. Lett.}, vol.~8, no.~5, pp. 2740--2747, 2023.

\bibitem{Li2022MSN}
B.~Li, Z.~Huang, T.~W. Chen, T.~Dai, Y.~Zang, W.~Xie, B.~Tian, and K.~Cai,
  ``Msn: Mapless short-range navigation based on time critical deep
  reinforcement learning,'' \emph{IEEE Trans. Intell. Transp. Syst.}, pp.
  1--10, 2022.

\bibitem{lecun2015deep}
Y.~LeCun, Y.~Bengio, and G.~Hinton, ``Deep learning,'' \emph{nature}, vol. 521,
  no. 7553, pp. 436--444, 2015.

\bibitem{Zhou2022Navigating}
Z.~Zhou, Z.~Zeng, L.~Lang, W.~Yao, H.~Lu, Z.~Zheng, and Z.~Zhou, ``Navigating
  robots in dynamic environment with deep reinforcement learning,'' \emph{IEEE
  Trans. Intell. Transp. Syst.}, vol.~23, no.~12, pp. 25\,201--25\,211, 2022.

\bibitem{Lim2020}
J.~Lim, S.~Ha, and J.~Choi, ``Prediction of reward functions for deep
  reinforcement learning via gaussian process regression,'' \emph{IEEE/ASME
  Trans. Mechatronics}, vol.~25, no.~4, pp. 1739--1746, 2020.

\bibitem{Wu2022}
K.~Wu, H.~Wang, M.~A. Esfahani, and S.~Yuan, ``Learn to navigate autonomously
  through deep reinforcement learning,'' \emph{IEEE Trans. Ind. Electron.},
  vol.~69, no.~5, pp. 5342--5352, 2022.

\bibitem{Tai2017}
L.~Tai, G.~Paolo, and M.~Liu, ``{Virtual-to-real deep reinforcement learning:
  Continuous control of mobile robots for mapless navigation},'' in \emph{IEEE
  Int. Conf. Intell. Robots Syst.}, vol. 2017-Septe, 2017, pp. 31--36.

\bibitem{shi2020}
H.~Shi, L.~Shi, M.~Xu, and K.-S. Hwang, ``End-to-end navigation strategy with
  deep reinforcement learning for mobile robots,'' \emph{IEEE Trans. Ind.
  Informat.}, vol.~16, no.~4, pp. 2393--2402, 2020.

\bibitem{xie2021}
L.~Xie, Y.~Miao, S.~Wang, P.~Blunsom, Z.~Wang, C.~Chen, A.~Markham, and
  N.~Trigoni, ``Learning with stochastic guidance for robot navigation,''
  \emph{IEEE Trans. Neural Netw. Learn. Syst.}, vol.~32, no.~1, pp. 166--176,
  2021.

\bibitem{Rana2023}
K.~Rana, V.~Dasagi, J.~Haviland, B.~Talbot, M.~Milford, and N.~Sünderhauf,
  ``Bayesian controller fusion: Leveraging control priors in deep reinforcement
  learning for robotics,'' \emph{Int. J. Rob. Res.}, vol.~42, no.~3, pp.
  123--146, 2023.

\bibitem{Jang2022hind}
Y.~Jang, J.~Baek, and S.~Han, ``Hindsight intermediate targets for mapless
  navigation with deep reinforcement learning,'' \emph{IEEE Trans. Ind.
  Electron.}, vol.~69, no.~11, pp. 11\,816--11\,825, 2022.

\bibitem{Leiva2020}
F.~Leiva and J.~Ruiz-del Solar, ``Robust rl-based map-less local planning:
  Using 2d point clouds as observations,'' \emph{IEEE Robot. Autom. Lett.},
  vol.~5, no.~4, pp. 5787--5794, 2020.

\bibitem{Pfeiffer2018}
M.~Pfeiffer, S.~Shukla, M.~Turchetta, C.~Cadena, A.~Krause, R.~Siegwart, and
  J.~Nieto, ``{Reinforced Imitation: Sample Efficient Deep Reinforcement
  Learning for Mapless Navigation by Leveraging Prior Demonstrations},''
  \emph{IEEE Robot. Autom. Lett.}, vol.~3, no.~4, pp. 4423--4430, 2018.

\bibitem{yu2020}
W.~Yu, J.~Tan, Y.~Bai, E.~Coumans, and S.~Ha, ``Learning fast adaptation with
  meta strategy optimization,'' \emph{IEEE Robot. Autom. Lett.}, vol.~5, no.~2,
  pp. 2950--2957, 2020.

\bibitem{Ali2021Bayesian}
A.~Ghadirzadeh, X.~Chen, P.~Poklukar, C.~Finn, M.~Björkman, and D.~Kragic,
  ``Bayesian meta-learning for few-shot policy adaptation across robotic
  platforms,'' in \emph{2021 IEEE/RSJ International Conference on Intelligent
  Robots and Systems (IROS)}, 2021, pp. 1274--1280.

\bibitem{Antonio2020}
A.~Loquercio, E.~Kaufmann, R.~Ranftl, A.~Dosovitskiy, V.~Koltun, and
  D.~Scaramuzza, ``Deep drone racing: From simulation to reality with domain
  randomization,'' \emph{IEEE Trans. Robot.}, vol.~36, no.~1, pp. 1--14, 2020.

\bibitem{Ishfaque2023}
A.~Ishfaque and B.~Kim, ``Real-time sound source localization in robots using
  fly ormia ochracea inspired mems directional microphone,'' \emph{IEEE Sens.
  Lett.}, vol.~7, no.~1, pp. 1--4, 2023.

\bibitem{Ding2022}
J.~Ding, Y.~Wang, H.~Si, S.~Gao, and J.~Xing, ``Three-dimensional indoor
  localization and tracking for mobile target based on wifi sensing,''
  \emph{IEEE Internet Things J.}, vol.~9, no.~21, pp. 21\,687--21\,701, 2022.

\bibitem{Haarnoja2018}
T.~Haarnoja, A.~Zhou, P.~Abbeel, and S.~Levine, ``{Soft actor-critic:
  Off-policy maximum entropy deep reinforcement learning with a stochastic
  actor},'' in \emph{35th International Conference on Machine Learning, ICML
  2018}, vol.~5, 2018, pp. 2976--2989.

\bibitem{de2021soft}
J.~C. de~Jesus, V.~A. Kich, A.~H. Kolling, R.~B. Grando, M.~A. d. S.~L.
  Cuadros, and D.~F.~T. Gamarra, ``Soft actor-critic for navigation of mobile
  robots,'' \emph{J. Intell. Robot. Syst.}, vol. 102, no.~2, p.~31, 2021.

\bibitem{Lillicrap2016}
T.~P. Lillicrap, J.~J. Hunt, A.~Pritzel, N.~Heess, T.~Erez, Y.~Tassa,
  D.~Silver, and D.~Wierstra, ``Continuous control with deep reinforcement
  learning,'' in \emph{4th International Conference on Learning
  Representations, {ICLR} 2016, San Juan, Puerto Rico, May 2-4, 2016,
  Conference Track Proceedings}, 2016.

\bibitem{zhang2022ipaprec}
W.~Zhang, Y.~Zhang, N.~Liu, K.~Ren, and P.~Wang, ``{IPAPRec: A promising tool
  for learning high-performance mapless navigation skills with deep
  reinforcement learning},'' \emph{{IEEE/ASME} Trans. Mechatronics}, vol.~27,
  no.~6, pp. 5451--5461, 2022.

\bibitem{liu2021regularization}
\BIBentryALTinterwordspacing
Z.~Liu, X.~Li, B.~Kang, and T.~Darrell, ``Regularization matters in policy
  optimization - an empirical study on continuous control,'' in
  \emph{International Conference on Learning Representations}, 2021. [Online].
  Available: \url{https://openreview.net/forum?id=yr1mzrH3IC}
\BIBentrySTDinterwordspacing

\bibitem{Bengio2009}
Y.~Bengio, J.~Louradour, R.~Collobert, and J.~Weston, ``Curriculum learning,''
  in \emph{Proceedings of the 26th Annual International Conference on Machine
  Learning}.\hskip 1em plus 0.5em minus 0.4em\relax Association for Computing
  Machinery, 2009, p. 41–48.

\bibitem{Fox1997}
D.~Fox, W.~Burgard, and S.~Thrun, ``The dynamic window approach to collision
  avoidance,'' \emph{IEEE Robot. Autom. Mag.}, vol.~4, no.~1, pp. 23--33, 1997.

\bibitem{vaughan2008massively}
R.~Vaughan, ``Massively multi-robot simulation in stage,'' \emph{Swarm
  Intell.}, vol.~2, pp. 189--208, 2008.

\bibitem{AMCL}
\BIBentryALTinterwordspacing
``{AMCL}.'' [Online]. Available: \url{http://wiki.ros.org/amcl}
\BIBentrySTDinterwordspacing

\end{thebibliography}

% biography section
% 
% If you have an EPS/PDF photo (graphicx package needed) extra braces are
% needed around the contents of the optional argument to biography to prevent
% the LaTeX parser from getting confused when it sees the complicated
% \includegraphics command within an optional argument. (You could create
% your own custom macro containing the \includegraphics command to make things
% simpler here.)
%\begin{IEEEbiography}[{\includegraphics[width=1in,height=1.25in,clip,keepaspectratio]{mshell}}]{Michael Shell}
% or if you just want to reserve a space for a photo:

% You can push biographies down or up by placing
% a \vfill before or after them. The appropriate
% use of \vfill depends on what kind of text is
% on the last page and whether or not the columns
% are being equalized.

%\vfill

% Can be used to pull up biographies so that the bottom of the last one
% is flush with the other column.
%\enlargethispage{-5in}
% that's all folks
\end{document}